%% file: main.tex
\documentclass[conference,compsoc]{IEEEtran}

\input{macros/math_commands}

\usepackage[utf8]{inputenc} 
\usepackage[T1]{fontenc}    
\usepackage{hyperref}       
\usepackage{url}            
\usepackage{booktabs}       
\usepackage{amsfonts}       
\usepackage{nicefrac}       
\usepackage{microtype}
\usepackage{algorithmic}
\usepackage{algorithm}
\usepackage{mathbbol}
\usepackage{tcolorbox}
\usepackage{url}
\usepackage{adjustbox}
\usepackage{multirow}
\usepackage{enumitem}
\usepackage{arydshln}
\usepackage{subcaption}
\usepackage{graphicx}
\usepackage{wrapfig}
\usepackage{soul} 

\input{macros/macros}

\graphicspath{{figures/}}

\title{Why Train More? Effective and Efficient Membership Inference via Memorization}
\author{Jihye Choi, Shruti Tople$^*$, Varun Chandrasekaran$^+$, Somesh Jha \\ $^*$ Microsoft, $^+$ University of Illinois Urbana-Champaign, University of Wisconsin-Madison}

\begin{document}

\maketitle

\input{sec_contents/0_abstract}
\input{sec_contents/1_intro}
\input{sec_contents/2_background_jihye}

\input{sec_contents/3_threat_model}

\input{sec_contents/4_motivation}

\input{sec_contents/5_theory_jihye}

\input{sec_contents/6_implementation}

\input{sec_contents/7_results}
\input{sec_contents/8_future}

\input{sec_contents/9_conclusions}


\newpage
\bibliographystyle{plain}
\bibliography{refs,refs_jihye}

\onecolumn
\appendices
\newpage
\input{sec_contents/appendix/theory}
\input{sec_contents/appendix/implementation}

\input{sec_contents/appendix/additional_results}

\end{document}

%% file: macros/math_commands.tex
\usepackage{amsmath,amsfonts,bm}

\newtheorem{thm}{Theorem}
\newtheorem{lemma}{Lemma}
\newtheorem{corollary}{Corollary}

\def\eqref#1{equation~\ref{#1}}

\def\1{\bm{1}}

\DeclareMathAlphabet{\mathsfit}{\encodingdefault}{\sfdefault}{m}{sl}
\SetMathAlphabet{\mathsfit}{bold}{\encodingdefault}{\sfdefault}{bx}{n}

\newcommand{\algcomment}[1]{\hfill {\color{blue} $\triangleright$ \emph{\small{#1}}}}

\newcommand{\mem}{\mathtt{mem}}

%% file: macros/macros.tex
\usepackage{xparse}
\usepackage{xspace}

\makeatletter
\def\@addpunct2#1{\ifnum\spacefactor>\@m \else#1\fi}
\newcommand{\para}[1]{\noindent\textbf{#1\unskip\@addpunct2{.}}~~}
\makeatother

\def\Dtr{{D^{\text{tr}}}}

\def\Dotr{{D_{\text{O}}^{\text{tr}}}}
\def\Dutr{{D_{\text{U}}^{\text{tr}}}}
\def\Dote{{D_{\text{O}}^{\text{te}}}}
\def\Dute{{D_{\text{U}}^{\text{te}}}}

\newcommand{\calA}{\mathcal{A}}
\newcommand{\calD}{\mathcal{D}}
\newcommand{\calX}{\mathcal{X}}

\newcommand{\fingerprint}[1][]{}
\NewDocumentCommand{\pw}{o}{
  \IfNoValueTF{#1}
    {\mathcal{P}}
    {\mathcal{P}(#1)}%
}

\newcommand{\mypara}[1]{\noindent\textbf{#1}}

\newcommand{\prf}{PoL\xspace}
\MakeRobust\prf

\newcommand{\pol}{PoL\xspace}
\MakeRobust\pol

\newcommand{\pols}{PoLs\xspace}
\MakeRobust\pols

\NewDocumentCommand{\oracle}{o}{
  \IfNoValueTF{#1}
    {Or}
    {Or(#1)}%
}
\NewDocumentCommand{\verifier}{o}{
  \IfNoValueTF{#1}
    {\mathcal{V}}
    {\mathcal{V}(#1)}%
}

\newcommand{\redst}[1][115pt]{\bgroup\markoverwith {\textcolor{red}{\makebox[0pt][l]{\rule[0.5ex]{#1}{0.4pt}}\rule[1ex]{#1}{0.4pt}}}\ULon}

\newcommand{\ie}{\textit{i.e.,}\@\xspace}
\newcommand{\eg}{\textit{e.g.,}\@\xspace}
\newcommand{\etal}{\textit{et al.}\@\xspace}

\newcommand{\jihye}[1]{\textcolor{blue}{JC: #1}}

\newcommand{\shruti}[1]{\textcolor{cyan}{ST: #1}}

\usepackage{arydshln}

\usepackage{tikz}
\usepackage{listofitems} 
\usetikzlibrary{arrows.meta} 
\usepackage[outline]{contour} 
\contourlength{1.4pt}

\usetikzlibrary{patterns}
\usetikzlibrary{decorations.markings}
\usetikzlibrary{positioning}
\usetikzlibrary{calc}
\usetikzlibrary{hobby}

\usepackage{dsfont}

\tikzset{>=latex} 
\usepackage{xcolor}
\colorlet{myred}{red!80!black}
\colorlet{myblue}{blue!80!black}
\colorlet{mygreen}{green!60!black}
\colorlet{myorange}{orange!70!red!60!black}
\colorlet{mydarkred}{red!30!black}
\colorlet{mydarkblue}{blue!40!black}
\colorlet{mydarkgreen}{green!30!black}
\tikzstyle{node}=[thick,circle,draw=myblue,minimum size=22,inner sep=0.5,outer sep=0.6]
\tikzstyle{node in}=[node,green!20!black,draw=mygreen!30!black,fill=mygreen!25]
\tikzstyle{node hidden}=[node,blue!20!black,draw=myblue!30!black,fill=myblue!20]
\tikzstyle{node convol}=[node,orange!20!black,draw=myorange!30!black,fill=myorange!20]
\tikzstyle{node out}=[node,red!20!black,draw=myred!30!black,fill=myred!20]
\tikzstyle{connect}=[thick,mydarkblue] 
\tikzstyle{connect arrow}=[-{Latex[length=4,width=3.5]},thick,mydarkblue,shorten <=0.5,shorten >=1]
\tikzset{ 
	node 1/.style={node in},
	node 2/.style={node hidden},
	node 3/.style={node out},
}

\newcommand{\cat}[1]{\;\operatorname{\big\|}\;}

\newcommand{\del}[1]{}

%% file: sec_contents/0_abstract.tex
\begin{abstract}

\textcolor{black}{
Membership Inference Attacks (MIAs) aim to identify specific data samples within the private training dataset of machine learning models, leading to serious privacy violations and other sophisticated threats. Many practical black-box MIAs require query access to the data distribution (the same distribution where the private data is drawn) to train \textit{shadow models}. By doing so, the adversary obtains models trained ``with'' or ``without'' samples drawn from the distribution, and analyzes the characteristics of the samples under consideration.
The adversary is often required to train more than hundreds of shadow models to extract the signals needed for MIAs; this becomes the computational overhead of MIAs. In this paper, we propose that by strategically choosing the samples, MI adversaries can maximize their attack success while minimizing the number of shadow models. First, our motivational experiments suggest memorization as the key property explaining disparate sample vulnerability to MIAs. We formalize this through a theoretical bound that connects MI advantage with memorization. Second, we show sample complexity bounds that connect the number of shadow models needed for MIAs with memorization.
Lastly, we confirm our theoretical arguments with comprehensive experiments; by utilizing samples with high memorization scores, the adversary can (a) significantly improve its efficacy regardless of the MIA used, and (b) reduce the number of shadow models by nearly two orders of magnitude compared to state-of-the-art approaches.
}

\end{abstract}

%% file: sec_contents/1_intro.tex
\section{Introduction}
\label{sec:intro}

Machine learning (ML) is ubiquitous; ML applications influence our lives in countless ways~\cite{petropoulos2020predicting,krizhevsky2010convolutional,ucci2019survey,openai2023gpt4}. Apart from clever algorithmic advances, the major proponent behind the rapid success of ML in the status quo is the availability of corpora of both structured and unstructured {\em data}~\cite{jain2020overview}. ML models are often trained on large volumes of data that is custom curated for the task at hand. Collecting and processing this data is both tedious and time-consuming, and requires significant investment. Additionally, the data used to train ML models is oftentimes sensitive, and protecting its privacy is paramount~\cite{choquette2021capc}. For instance, ML models trained for medical applications involve health-related information pertaining to patients; the privacy of this data is protected by regulations such as HIPAA~\cite{annas2003hipaa}.

The widely held belief was that, since the data is processed in non-intuitive ways to obtain an ML model, purely interacting with the model would not leak information about the data used to train it. However, prior work by Fredrikson \etal~\cite{fredrikson2015model} demonstrated how an adversarial entity could {\em reconstruct} the training data given query access to the trained model. This was the first attack on the privacy of the training data. 
Subsequent work, dubbed {\em membership inference} (MI), is a weaker decision variant; it aims to infer if a particular record was present (or absent) in the training data given access to a trained model. 
Follow-up research aims to determine the {\em advantage} of such an MI adversary \ie how likely is the adversary to succeed in their attempts, mostly based on connections to differential privacy~\cite{mahloujifar2022optimal,sablayrolles2019white,yeom2018privacy}. 
Recent work by Salem \etal~\cite{salem2022sok} highlights how understanding advantage in the context of MI is paramount to understanding more sophisticated privacy threats. 

In this paper, we focus on the fact that such MI advantage (\ie adversary success) widely varies across different data samples, and study how the adversary can identify "more vulnerable" data to employ more successful/accurate and more (computationally) efficient MI attacks (MIAs).

\noindent{\bf Contributions:} Recent research by Feldman \etal~\cite{feldman2020longtail,feldman2020neural} and Brown \etal~\cite{brown2021memorization} formalizes {\em label memorization}, which explains how overparameterized ML models (such as deep neural networks) need to memorize the label of certain atypical samples (that belong to low-density subsets in the data distribution) to facilitate good generalization behaviors. 
To this extent, our main contribution here is to view the MIA framework through the lens of memorization, and provide the {\em first formal bound} of MI advantage as a function of label memorization. 
This bound is particularly interesting given that it unearths the characteristics of samples for which MIAs are most successful, as label memorization is a data-dependent phenomenon.

Furthermore, we establish a connection between label memorization and the computational efficiency of MIAs.
It is worth noting that many MIAs necessitate the training of a substantial number of \textit{shadow models}, thereby creating a computational bottleneck~\cite{shokri2017membership,song2021systematic, watson2021importance, ye2021enhanced}.
These shadow models serve to simulate and analyze the statistical patterns exhibited by models trained with or without data samples under consideration.
By framing MIA as a hypothesis testing problem, akin to previous works~\cite{carlini2022membership, ye2021enhanced}), we note that the sample complexity of hypothesis testing corresponds to the number of shadow models. We then provide a formal bound on the sample complexity in terms of label memorization, implying that MIAs targeting highly memorized samples necessitate fewer shadow models to achieve successful results with a high level of confidence.




Lastly, we conduct thorough extensive evaluation over multiple vision datasets and models, to support our argument that our explanation of disparate impact yields both (a) more efficient, and (b) computationally inexpensive MIAs compared to state-of-the-art approaches. 
In particular, with highly memorized data, we are able to achieve AUROC of $0.99$ and TPR of $100 \%$ at $0.1\%$ FPR with less than $20$ shadow models, whereas achieving the same success with random data often requires more than $200$ shadow models. 


\noindent{\bf Comparison to Prior Work:} Previously, Kulynych \etal~\cite{kulynych2019disparate} observe that adversarial entities are more successful if they are subgroup-aware. This implicitly suggests that the property of interest in the context of their study is sub-population susceptibility. They formalize the notion of disparate vulnerability, which highlights the lack of generalization among different subgroups, and exploit this for their attacks. 
Yeom \etal~\cite{yeom2018privacy} were the first to study the impact of feature influence in a black-box setting. They notice that certain features are more important than others, and this can be exploited for privacy attacks. 
Leino \etal~\cite{leino2020stolen} build atop the observations made by Yeom \etal; they notice that ML models behave differently (in terms of predictive accuracy) when inputs contain specific features, and this is exploited to launch more effective attacks in a white-box setting. 
The recent attack by Carlini \etal~\cite{carlini2022privacy} is a composition of memorization estimation and hypothesis testing, and provide a brief demonstration of their disparate performance on in-distribution vs. out-of-distribution (OOD) samples (\ie samples drawn from the different distribution than usual training data). 
However, we present memorization as the "key" property to explain disparate sample susceptibility to MIAs. We also make {\em the first formal characterizations} of MI advantage as a function of memorization.

In summary, our main contributions include:
\begin{enumerate}
\itemsep0em
\item A new explanation for the disparate impact of MIAs based on memorization. In particular, we see how MIAs are more effective for samples that are highly likely to be memorized (refer \S~\ref{sec:motivation}).
\item 
Theoretical underpinning for the above explanation, including a novel lower bound on MIA advantage based on label memorization. This lower bound establishes the formal connections between privacy attacks and label memorization (see \S~\ref{sec:adv-mem}).
\item Sample complexity results that offer insights into optimizing the efficiency of MIAs employing shadow models, particularly for highly memorized samples. We show that the number of shadow models required is inversely proportional to the memorization score of a sample, meaning higher memorization scores demand fewer models (see \S~\ref{sec:sample-mem}).
\item 
Extensive experimental evaluation validating the theoretical constructs across various vision datasets and model architectures. 
Our experiments further illustrate that any MIA proves more effective for highly memorized samples compared to random OOD samples or samples from under-represented subpopulations in the data. This holds true even when the adversary aims for high true positive rates under low false positive rates (see \S~\ref{sec:results}).
\end{enumerate}

%% file: sec_contents/2_background_jihye.tex
\section{Background}
\label{sec:background}

In this section, we introduce the background and formalism that we use throughout this paper. 

\subsection{Machine Learning}
\label{subsec:ml}

\mypara{Supervised Machine Learning:} Distribution $\mathcal{D}$ captures the space of inputs and outputs, from which a dataset $S$ of size $s$ is sampled (\ie $S \sim \mathcal{D}^s$) such that $S = \{z_i\}_{i=1}^s$. Each $z_i$ is of the form $(x_i,y_i)$, where $x_i \in \mathcal{X}$ is the space of inputs (\eg images) and $y_i \in \mathcal{Y}$ is the space of outputs (\eg labels). Using this dataset, a learning algorithm $L$ (\eg stochastic gradient descent~\cite{bottou2012stochastic}) can create a trained model \ie $\theta \sim L(S)$ by minimizing a suitable objective $\mathcal{L}$ (\eg cross-entropy loss~\cite{zhang2018generalized}). 

\mypara{Membership Inference:} For many applications, ML models are trained on large corpora of data, some of which is sensitive and private~\cite{shokri2017membership}. MI is a class of attacks on the privacy of the data used to train the model, and are training distribution-aware~\cite{yeom2018privacy,shokri2017membership,carlini2022membership,jayaraman2020revisiting,song2021systematic,murakonda2021quantifying,ye2021enhanced,sablayrolles2019white,leino2020stolen}. Adversaries aim to infer if a sample was present in the training dataset, given access to a trained model. They often do so by checking for effects due to the presence or absence of the point under consideration on a large set of shadow models -- those that are trained on varying data subsets to estimate these effects. MIAs were used to test if data is successfully deleted from ML models~\cite{bourtoule2021machine}, with some caveats~\cite{kong2023can}. Data deletion itself is susceptible to MIAs~\cite{chen2021machine}. 



\mypara{Label Memorization:} ML models are vastly overparameterized, and are known to fit to even random labels~\cite{feldman2020longtail}. 
This is termed \textit{memorization}. For a sample $z_i$, given dataset $S$ (without $z_i$), label memorization is formalized as follows:
\begin{equation}
\label{eq:memorization}
\mem(L,S,z_i) := \Pr_{\theta \sim L(S^{(i)})}[\theta(x_i) = y_i] - \Pr_{\theta \sim L(S)}[\theta(x_i) = y_i]   
\end{equation}
where $S^{(i)} = S \cup \{z_i\}$. We term the value of Equation~\ref{eq:memorization} as a memorization score.


\subsection{Hypothesis Testing}
\label{subsec:hypo_testing}

A hypothesis test is a method of statistical inference where one checks if the data at hand is sufficient to support a particular hypothesis. Let $P_0$ and $P_1$ denote two distributions with support $\mathcal{X}$. A hypothesis test $T: \mathcal{X}^\star \rightarrow \{ 0,1 \}$ takes a sequence of $n$ elements $\sigma = \{ o_1,\cdots, o_n \}$ and predicts $0$ (\ie $\sigma$ is generated by $P_0$) or $1$ (\ie $\sigma$ is generated by $P_1$). There are two types of errors associated with hypothesis testing: 
\begin{itemize}
\itemsep0em
\item {\bf Type I error}: The probability that the test outputs $P_1$ if $P_0$ is true \ie $\Pr_{\sigma \sim P_0^n} ( T(\sigma) = 1)$.
\item {\bf Type II error}: The probability that the test outputs $P_0$ if $P_1$ is true \ie $\Pr_{\sigma \sim P_1^n} (T(\sigma) = 0)$. \end{itemize}
The ideal goal of a hypothesis test is to achieve Type I error below an application specific threshold. 

\noindent The advantage of a test $T$ is defined as,
\begin{equation}
\label{eq:adv_HT}
    Adv_n (T) = \Pr_{\sigma \sim P_0^n} ( T(\sigma) = 0) - \Pr_{\sigma \sim P_1^n} (T(\sigma) = 0)
\end{equation}


\mypara{Log-Likelihood Ratio (LLR) Test:} 
The Neyman-Pearson lemma~\cite{neyman1933ix} states that the LLR test achieves the best Type II error for a given bound on the Type I error, and is optimal. Formally,
\begin{equation}
\label{eq:llr}
    \text{LLR}(\sigma) = \log \frac{P^n_0(\sigma)}{P^n_1(\sigma)} = \log \prod_{i=1}^n \frac{P_0(o_i)}{P_1(o_i)}
\end{equation}
If $\text{LLR}(\sigma) \geq \kappa$ (for some constant $\kappa$), the test outputs $P_0$, and otherwise outputs $P_1$. Consider the soft version of the above LLR test, $\text{sLLR}(x)$, which outputs $0$ with probability $g(x)$ and $1$ with probability $1-g(x)$, where $g(x)$ is:
\begin{equation*}
\frac{e(P_0,P_1)}{1+e(P_0,P_1)}     
\end{equation*}
where $e(P_0,P_1)$ denotes $\exp(\frac{1}{2} \log \frac{P_0 (o)}{P_1 (o)} )$. Cannone \etal~\cite{canonne2019structure} describe the advantage of the sLLR test using a single sample as follows:
\begin{tcolorbox}
\begin{lemma}
\label{lem:sLLR}
For any two distributions $P_0, P_1$, the advantage of $\text{sLLR}$ test with $n=1$ is $Adv_1(\text{sLLR}) = H^2 (P_0, P_1)$.
\end{lemma}
\end{tcolorbox}

\mypara{Hellinger distance:}
The square of the Hellinger distance between two distributions $P_0, P_1$ over measure space $\mathcal{X}$ is defined as
\begin{equation}
    H^2(P_0, P_1) = \frac{1}{2} \int_\mathcal{X} (\sqrt{P_0} - \sqrt{P_1}) ^2 
\end{equation}
which captures the ``distance'' between two probability distributions.

\mypara{Sample Complexity:}
{\it Sample complexity} $M_\alpha^{P_0, P1} (T)$ of a test $T$ is defined such that for all $n \geq M_\alpha^{P_0,P_1} (T)$, the advantage $Adv_n (T) \geq \alpha$. The sample complexity of the optimal test $T^\star$ serves as the lower bound for the number of samples necessary to achieve an error probability $1-\alpha$:
\begin{equation*}
    M_\alpha^{P_0, P1} (T^\star) = \min_{T} ~ M_\alpha^{P_0, P1} (T)
\end{equation*}
where $T^\star$ corresponds to LLR test in Equation~\ref{eq:llr} by Neyman-Pearson lemma.

\begin{tcolorbox}
\begin{lemma}
\label{lem:opt_sample}
The optimal sample complexity of a test to disambiguate between $P_0$ and $P_1$ is $\Theta (\frac{1}{H^2(P_0, P_1)})$.
\end{lemma}
\end{tcolorbox}

This bound is tight (\ie both lower-bounded and upper-bounded)~\cite{bar2002complexity,canonne2019structure}.




%% file: sec_contents/3_threat_model.tex
\section{Problem Overview}
\label{sec:threat}

\subsection{Threat Model} 
\label{sec:threat}

\mypara{Membership Inference Game:} The $MI$ game defines the interaction between two parties: an adversary $\calA$ aiming to perform MI, and a challenger $\mathcal{C}$ that is responsible for training. We assume $\calA$ knows the data distribution $\calD$ and the learning algorithm $L$ but not the randomness used by $L$. The output of $L$ belongs to the set $\Theta$. 
\begin{enumerate}
\itemsep0em 
    \item $\mathcal{C}$ picks a dataset $S$ sampled according to $\calD$. 
    \item $\calA$ picks $z \in \calD$ and sends it to $\mathcal{C}$.
    \item $\mathcal{C}$ picks a random bit $b_{\mathcal{C}} \leftarrow \{ 0,1 \}$.  If $b_{\mathcal{C}}=0$, let $S' = S$; otherwise $S'= S \cup \{ z \}$. $\mathcal{C}$ executes $L$ on $S'$ and sends the result $\theta_L \in \Theta$ to $\calA$.
    \item $\calA$ guesses a bit $b_{\calA}$. If $b_{\mathcal{C}}= b_{\calA}$, then the output of the game is $1$ (indicating that the adversary won the game); otherwise the output of the game is $0$ (indicating that the adversary lost the game).
\end{enumerate}
The advantage of the adversary is written as the difference between $\calA$’s true and false positive rates as follows~\cite{yeom2018privacy},
\begin{equation}
\label{eq:adv_MI}
\mbox{Adv}  (L,\calA) ~=~ \Pr(b_{\calA} = 0 \mid b_{\mathcal{C}} = 0) - \Pr(b_{\calA}=0 \mid b_{\mathcal{C}}=1)
\end{equation}
Let $C_{\calA}$ be a class of adversaries (\eg probabilistic polynomial time algorithm adversaries). We can define 
$\mbox{Adv} (L,C_{\calA})$ as $\sup_{\calA \in C_{\calA}} \mbox{Adv}  (L,\calA)$. When the class $C_{\calA}$ is implicit from the context we still write it as $\mbox{Adv} (L,\calA)$.

The random variable corresponding to the $MI$ game is written as 
$MI_{L,\calA}$. We define $MI^s_{L,\calA}$ separately to denote the \textit{stronger game} where the dataset $S$ is also picked by $\calA$~\cite{mahloujifar2022optimal} (in step 1). In our work, such stronger game is assumed. \textcolor{black}{Such an assumption is not far-fetched. In scenarios where trusted auditors verify data deletion through MIAs~\cite{ma2022learn}, the auditors (in this case, the adversary) are assumed to know both the dataset $S$ and the point being deleted $z$. Additionally, such a strong assumption allows us to formally prove the connection between MIAs and memorization against the worst-case adversary.}


Carlini \etal~\cite{carlini2022membership} argue that an effective MIA is the one that {\em reliably} identifies few members of a sensitive dataset; such an attack is more preferred than those that achieve high aggregate success but are unreliable. They argue that this is how conventional security applications are evaluated~\cite{ho2017detecting}. We follow the same setup. Our privacy adversary wants to evaluate their MIA's (formalized as algorithm $\mathcal{A}$) true-positive rate (TPR) at low false-positive rates (FPR); success implies high TPR at low FPR regimes.


\subsection{Problem Statement}

As noted in \S~\ref{sec:intro}, prior work has attributed the disparate impact of MIA to the lack of generalization (or overfitting)~\cite{yeom2018privacy}, or have noticed that MIAs are more successful on outliers~\cite{carlini2022membership} and under-represented subpopulations~\cite{kulynych2019disparate}. However, most MIAs that are designed based on these observations are not always effective (as we will describe in \S~\ref{sec:motivation}); not {\em all samples} that are identified by these observations are highly susceptible to MIAs. 
To this end, we address the following two questions:
\begin{enumerate}
\itemsep0em
\item[{\bf Q1.}] What property of data best explains the disparate performance of MIAs?
\item[{\bf Q2.}] Can the MI adversary leverage this property to launch more effective and efficient attacks?
\end{enumerate}
To answer the first question (\textbf{Q1}), we first taxonomize the aforementioned phenomena (explaining disparate performance) and provide strong motivational (empirical) evidence that memorization is the most appropriate explanation for data's disparate susceptibility to MIAs (\S~\ref{sec:motivation}). 
Then, we address the second question (\textbf{Q2}) by establishing a theoretical connection between game $MI^s_{L,\calA}$ and label memorization. We demonstrate that by utilizing highly memorized data samples, the adversary can significantly improve both efficacy and computational efficiency (\S~\ref{sec:theory}). Finally, we provide thorough empirical results to support our theoretical findings (\S~\ref{sec:implementation} and \S~\ref{sec:results}).  



\mypara{Our Main Thesis:} We study a setting where the MI adversary can strategize its game by choosing samples that are highly likely to be memorized (\ie a large value in Equation~\ref{eq:memorization}) to add in game $MI^s_{L,\calA}$. 
Here, we assume the existence of a memorization oracle $\mathcal{O}_{mem}$ that returns an approximate memorization score for any sample chosen by the adversary. These samples may come from the same distribution $\mathcal{D}$ that was used to obtain $S$, or can come from a different distribution (refer \S~4 of Carlini \etal~\cite{carlini2021extracting}) \textcolor{black}{In our experiments in \S~\ref{sec:results}, data points $z$ are sampled from a different distribution than $S$, and the memorizaton oracle is instantiated by running the algorithm proposed by Feldman and Zhang~\cite{feldman2020neural}}. The presence (or absence) of such points leads to a characteristic signal which the adversary exploits to launch MIAs. 
We would like to stress that the notion of (label) memorization is one that is dependent only on the dataset; any well-generalized model learnt using this dataset will show similar memorization trends among samples (if not the exact same values).

\textcolor{black}{More formally, we wish to prove the following:} \textcolor{black}{Assume a strong adversary $\calA$ that has complete knowledge of the dataset $S$ used to train a machine learning model using algorithm $L$.}
\begin{itemize}
\item \textcolor{black}{For a sample $z^*$, $\calA$ can successfully\footnote{For success defined in \S~\ref{sec:threat}.} identify its membership in $S$ if $\mem(L,S,z^*)=1$ \ie $\mbox{Adv}  (L,\calA) = 1$.}
\item \textcolor{black}{If $\calA$ utilizes $m$ shadow models for launching an MIA for a sample $z$, it requires $m^* \leq m$ shadow models if $\mem(L,S,z^*)=1$.}
\end{itemize}

\input{sec_contents/ood_table_figs}


%% file: sec_contents/ood_table_figs.tex
\begin{table*}[bht]
\resizebox{\textwidth}{!}{
\centering
\footnotesize
\begin{tabular}{ll|ll|ccccc}
\toprule
\multirow{2}{*}{\bf{Over-represented}} & \multirow{2}{*}{\bf{Under-represented}} & \multirow{2}{*}{\bf $\alpha$} & \multirow{2}{*}{\bf \# Samples} & \multirow{2}{*}{\bf Model} & \multicolumn{2}{c}{\bf Train Acc. (\%)} & \multicolumn{2}{c}{\bf Test Acc. (\%)} \\ \cline{6-9}
 & & & & & $\Dotr$ & $\Dutr$ & $\Dote$ & $\Dute$ \\
\hline \hline
\multirow{4}{*}{MNIST} & None & 1 & 0 & CNN32 & 98.54 $\pm$ 0.08 & N/A & 98.36 $\pm$ 0.14 & N/A\\
& augmented MNIST & 0.99 & 1000 & CNN32 & 99.15 $\pm$ 0.13  & 89.10 $\pm$ 1.7 & 98.43 $\pm$ 0.19 & 65.88 $\pm$ 1.39 \\
& SVHN & 0.99 & 1000 & CNN32 & 99.44 $\pm$ 0.05 & 90.07 $\pm$ 0.66 & 98.60 $\pm$ 0.09 & 64.76 $\pm$ 1.31 \\
&CIFAR-10 & 0.93 & 6000 & CNN64 & 99.27 $\pm$ 0.86 & 82.69 $\pm$ 6.84 & 98.18 $\pm$ 0.47 & 40.72 $\pm$ 1.17 \\ 
\cdashline{1-9}

 \multirow{3}{*}{CIFAR-10} & None & 1 & 0 & ResNet-9 & 97.06 $\pm$ 0.77 & N/A & 90.02 $\pm$ 0.02 & N/A \\
 &augmented CIFAR-10 & 0.98 & 1000 & ResNet-18 & 96.39 $\pm$ 0.36 & 86.85 $\pm$ 2.14 & 87.79 $\pm$ 1.16 & 54.41 $\pm$ 1.79 \\
 & CIFAR-100 & 0.98 & 1000 & ResNet-9 & 96.59 $\pm$ 0.21 & 79.12 $\pm$ 21.7 & 88.39 $\pm$ 0.59 & 35.37 $\pm$ 2.55 \\
 \cline{1-9}
\end{tabular}
}
\caption{\textbf{Salient features of the datasets used in our experiments.} Distance represents the distance between $\Dotr$ and $\Dutr$ (larger is more). Accuracy is reported in $95\%$ confidence interval -- for each iteration in bootstrapping, we include $70\%$ of $\Dtr$ as the training set, and repeat $2000$ times (as done by Feldman~\cite{feldman2020neural}). We also ensure that each sample in $\Dtr$ appears in the training set for half of the $2000$ iterations, following the setup of Carlini \etal~\cite{carlini2022membership}.}
\label{tab:salient}
\end{table*}

\begin{figure*}[htb]
\centering
\begin{subfigure}{\textwidth}
\includegraphics[width=0.32\textwidth]{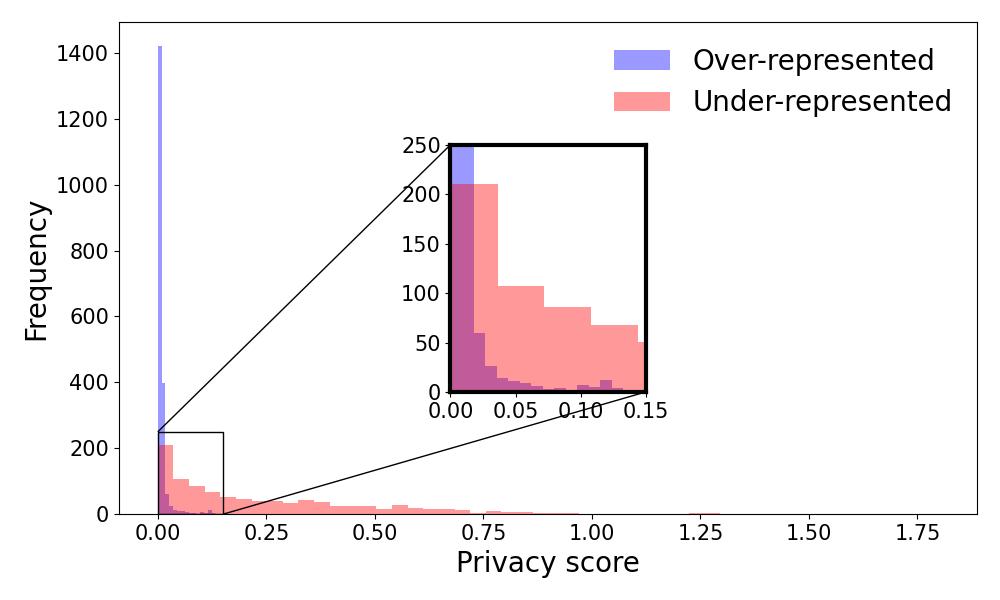}
\includegraphics[width=0.32\textwidth]{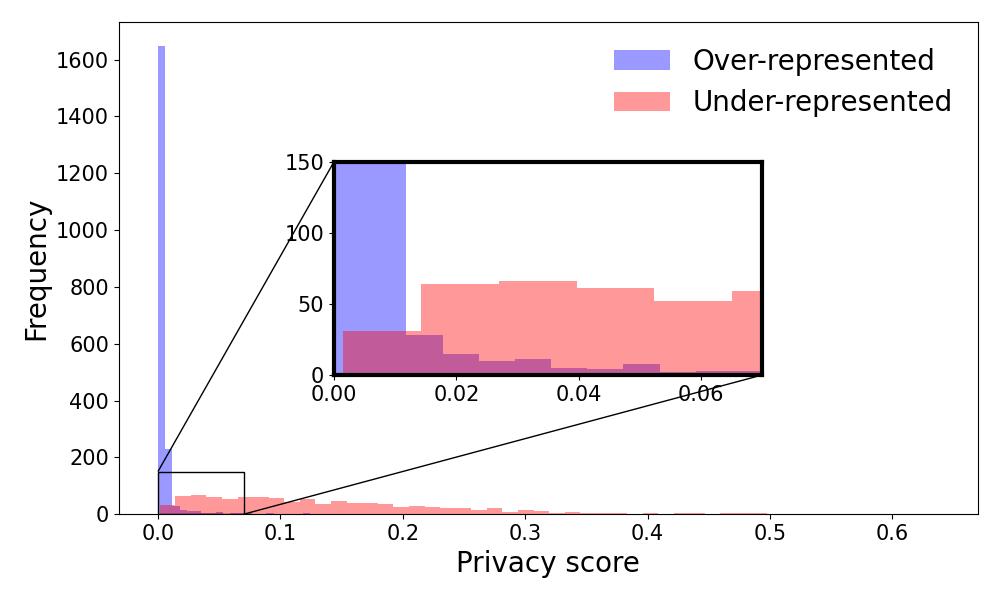}
\includegraphics[width=0.32\textwidth]{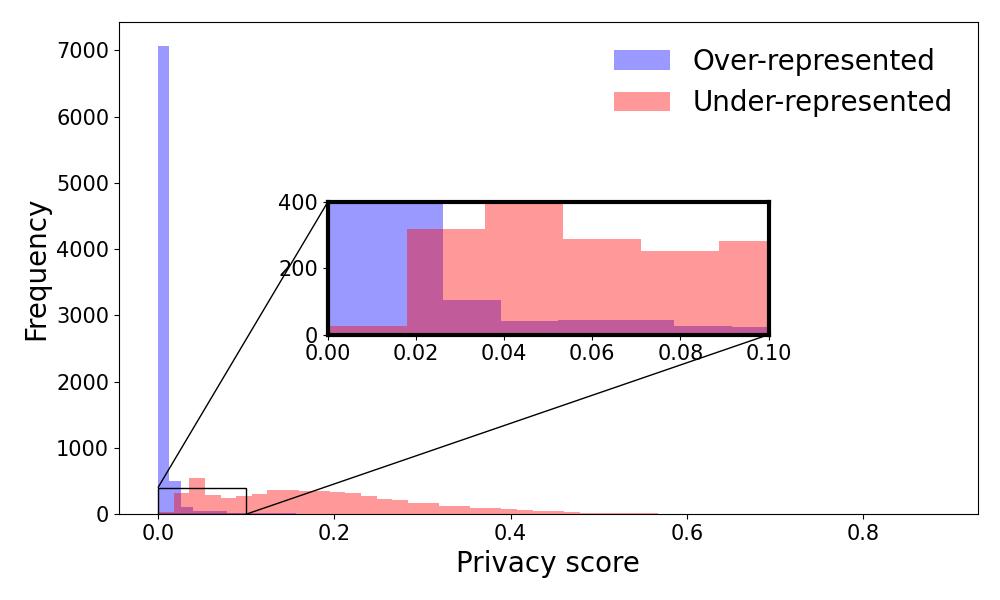}
\caption{Distribution of privacy scores between over-represented group (blue) vs. under-represented group (red)}
\label{fig:privacy-distr-mnist}
\end{subfigure}
\\
\vspace{2mm}
\centering
\begin{subfigure}{\textwidth}
\includegraphics[width=0.32\textwidth]{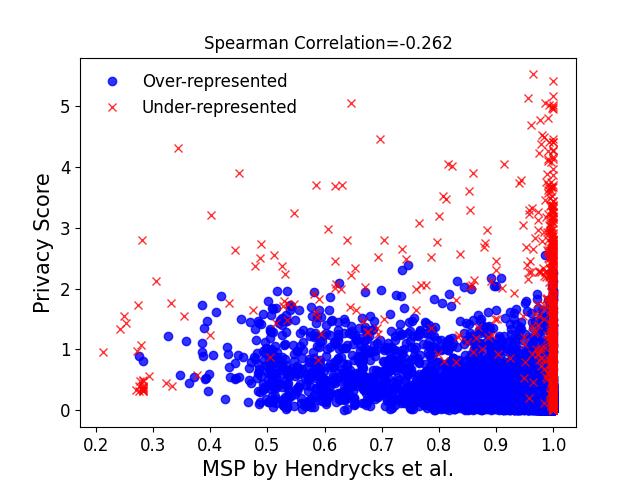}
\includegraphics[width=0.32\textwidth]{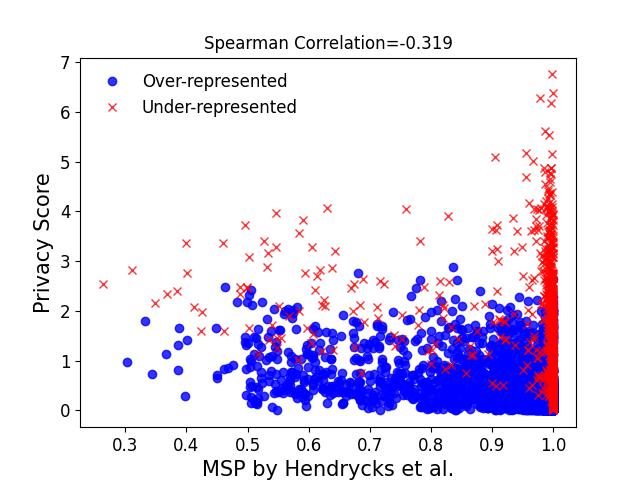}
\includegraphics[width=0.32\textwidth]{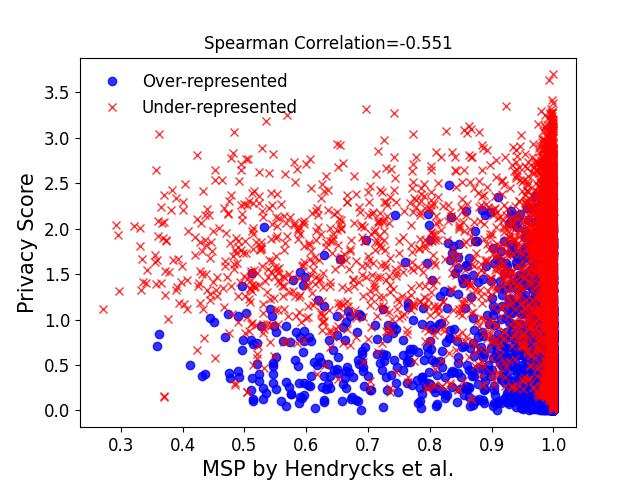}
\caption{OOD score from MSP detector~\cite{hendrycks2016msp} (on the x-axis) vs. privacy score (on the y-axis).}
\label{fig:msp-privacy-mnist}
\end{subfigure}

\caption{\textbf{OOD data is not always more susceptible to MI attack than ID data.} (i) \textbf{L}: CNN32 trained on MNIST+augMNIST  ({\bf V1}), (ii) \textbf{M}: CNN32 trained on MNIST+SVHN ({\bf V2}), (iii) \textbf{R}: CNN64 trained on MNIST+CIFAR-10 ({\bf V3}).
}
\label{fig:privacy-ood-mnist}
\end{figure*}

%% file: sec_contents/4_motivation.tex
\section{Motivational Experiments}
\label{sec:motivation}




In this section, we aim to understand what properties of data explain the disparate performance of MIAs. This can be leveraged by the adversary to employ more successful attacks (later in \S~\ref{sec:theory} and \S~\ref{sec:results}).
Prior work shows that the disparate performance is associated with how atypical the sample is w.r.t other points in the training dataset. We wish to understand if this theory provides a nuanced (and holistic) understanding of MIA success with in-depth experiments.

\noindent{\bf Experimental Setup:} We consider a training data $\Dtr$ consisting of two different datasets (say $\Dotr$ and $\Dutr$) with a mixing ratio $\alpha \in [0,1]$ (\ie to obtain $\Dtr = \alpha \cdot \Dotr + (1 - \alpha) \cdot \Dutr$).  In such settings, one dataset is an {\em over-represented group} (\ie $\Dotr$ when the mixing coefficient is greater than 0.5), and the other is an {\em under-represented group} (\ie $\Dutr$). 
\textcolor{black}{Such composite datasets have been recognized in many contexts; for instance, in federated learning, data in many settings, data is pooled from differet sources such as devices from various geographic regions~\cite{li2020federated}; modern datasets are long-tailed and composed of varied subpopulations~\cite{zhu2014longtail1,van2017longtail2,babbar2019longtail3}.
}

\textcolor{black}{\mypara{Our notion of Atypicality:} We realize the under-represented group by picking data samples "far" from the over-represented group, which is commonly referred to as out-of-distribution (OOD) data. Commonly, this is realized by considering \emph{covariate-shifted} OOD data (\ie data generated by applying semantics-preserving transformations to the training data, preserving the original labels), or \emph{semantic} OOD data (\ie data having completely disjoint label sets from the normal training data)~\cite{bai2023scone}.}

In our experiment, the over-represented group is fixed as the entire dataset of MNIST~\cite{lecun2010mnist} (60,000 labeled samples), while the choice of under-represented group varies. We consider three variants: ({\bf V1}) MNIST augmented by the approach proposed by Hendrycks \etal~\cite{hendrycks2022pixmix} (henceforth referred to as augMNIST), ({\bf V2}) cropped SVHN~\cite{svhn}, and ({\bf V3}) CIFAR-10~\cite{cifar10}. 
\textcolor{black}{
{\bf V1} and {\bf V2} correspond to covariate-shifted OOD, while {\bf V3} is semantic OOD.
}
To construct $\Dutr$ (for a given variant), we choose a subset of samples (more details follow) from the corresponding training set; care is taken to ensure the consistent label assignment to both $\Dotr$ and $\Dutr$ \footnote{Consistent and balanced label assignment across classes; \ie all "airplane" images in CIFAR-10 are labeled as class 0 along with digit "$0$" images of MNIST etc.}. All samples are (a) resized to match the dimension of MNIST samples ($28 \times 28 \times 1$), and (b) converted to grayscale\footnote{Gray Image = $0.2989 \times \text{R} + 0.5870 \times \text{G} + 0.1140 \times \text{B}$.}. We use CNN models from the work of Carlini \etal~\cite{carlini2022membership} (for more detailed descriptions, refer to Appendix~\ref{app:implementation-details}).

\subsection{Are All OOD Samples Susceptible?} 
\label{subsec:connect_ood}

In this subsection, we aim to examine if the OOD-ness of data is a sufficient explanation for susceptibility to MIAs.
To check if OOD samples are more susceptible to MIAs, we choose a random subset from each of the aforementioned variants ({\bf V1-3}) as our under-represented group ($\Dutr$) and train models to achieve reasonable training and test accuracy (greater than 80\% for train, and greater than 50\% in most cases for test) on both groups. To do so, we need at least 1000 samples for augmented MNIST and cropped SVHN, and at least 6000 CIFAR-10 samples (10\% of the total dataset); all these samples are chosen such that there are equal number of samples for each class. Complete statistics (including the mixing ratio) are presented in Table~\ref{tab:salient}. 

In all the experiments, we first compute the privacy score which captures the vulnerability of a sample to an MIA (higher is more vulnerable)~\footnote{\textcolor{black}{Privacy score = $\frac{|\mu_{\text{in}} - \mu_{\text{out}}|
}{\sigma_{\text{in}} + \sigma_{\text{out}}}$ where $\mu$ and $\sigma$ are statistical parameters to characterize when a data sample is included (or excluded) in the training set. Refer to \S~VII-B of Carlini \etal~\cite{carlini2022membership} for the full description.
}}
We then compare the distributional pattern of privacy scores between over-represented and under-represented groups (blue and red histograms, respectively, in Figure~\ref{fig:privacy-distr-mnist}).
We also measure the correlation between the privacy score and OOD-ness of data--quantified via the output score from OOD detectors. Here we consider the two representative OOD detectors; MSP detector by Hendrycks \etal~\cite{hendrycks2016msp} (see Figure~\ref{fig:msp-privacy-mnist}) and Energy detector by Liu \etal~\cite{liu2020energy} (see Figure~\ref{fig:energy-privacy-mnist} in Appendix~\ref{app:additional-motivating-results}). Higher value implies more OOD-ness.

\noindent{\bf Observations:} Several important observations can be made. First, we want to understand if {\em all} OOD samples are highly susceptible to our MIA. This is not the case. 
In Figure~\ref{fig:privacy-distr-mnist}, the majority of under-representated data (red histogram) has higher privacy score than over-represented data (blue histogram), but they are not completely separated from each other. 
Accordingly, the red points (representing under-represented data) and blue points (representing over-represented data) are overlapped in Figure~\ref{fig:msp-privacy-mnist}, and a large fraction of the under-represented samples are not highly susceptible to MIAs (\ie many OOD samples have the privacy score below 2 on the y-axis). 
Next, we want to examine the degree of correlation between being OOD and MIA susceptibility. We capture correlations using the Spearman's rank correlation coefficient with $p$-value less than $0.001$ for statistical significance; (absolute) values less than 0.4 are considered to represent a weak correlation, values between 0.4 and 0.59 are considered to represent a moderate correlation, and values greater than 0.6 are considered to represent a strong correlation; the sign dictates whether the correlation is positive or not (irrelevant for our discussion). 
We observe that the absolute degree of correlation below 0.4 or at most 0.551 (see the numbers on the subfigure titles in Figure~\ref{fig:msp-privacy-mnist}, meaning that there is {\em no strong correlation} between OOD scores (measured by the approach of Hendrycks \etal~\cite{hendrycks2016msp}) and MIA success; experiments with other OOD detector~\cite{liu2020energy} show similar trends (see Figure.~\ref{fig:energy-privacy-mnist} in Appendix~\ref{app:additional-motivating-results}). 

\noindent{\bf Implications:} The aforementioned observations strongly indicate that relying on OOD scores to detect membership status is not a definitive approach~\footnote{We utilize popular baselines from literature to capture this effect. We acknowledge that there are no perfect measures to determine OOD-ness, the OOD detector outputs are mere proxies of them.}. Among all OOD data points, only a subset of them exhibits high privacy scores (\ie more susceptibility to MI attacks).
This suggests that while previous explanations on the disparate impact of MI vulnerability hold true to some extent, they are rather coarse and require further refinement. 
We posit that this is because the conventional definitions of OOD-ness found in the literature (\eg label disjointedness) are not precisely aligned with the correct characteristics of samples that semantically differentiate them from the majority of the data, thereby failing to explain their vulnerability to MIAs. In fact, current definitions of OOD-ness often suffer from flaws~\cite{bitterwolf2023or}, and the results obtained based on such assumptions lack reliability and precision.

\input{sec_contents/mem_tab_figs}
\subsection{Are Memorized Samples Susceptible?}
\label{subsec:connect_memo}

A second phenomenon that can cause disparate performance of MIAs is the model's capability of memorizing a sample (irrespective of it being from within or outside the training data distribution). Brown \etal~\cite{brown2021memorization} note that the mutual information between a model's parameters and samples that are memorized is high; this leads us to believe that such samples are highly likely to be susceptible to MIAs. We proceed to identify the samples with high label memorization scores ($ > 0.8$), calculated based on the definition in \S~\ref{subsec:ml}~\footnote{For the complete description of estimating privacy and memorization scores, see Algorithm~\ref{alg:mem-privacy} in Appendix~\ref{app:algorithm}.}.
Note that by doing so, we reduce the size of the under-represented group (details in Table~\ref{tab:salient-singleton} in Appendix~\ref{app:additional-motivating-results}). We then measure the Spearman's rank correlation coefficient between the memorization score and the privacy score.  

\noindent{\bf Observations:} Several observations can be made from Figure~\ref{fig:privacy-mem-singleton_top}. First, the samples that have the high memorization scores (\ie the red points) roughly correspond to those samples which have a high privacy score in Figure~\ref{fig:msp-privacy-mnist}; this suggests that the OOD samples with high memorization scores are indeed those that are highly susceptible to MIAs.  Next, we visualize the samples corresponding to different memorization scores from each dataset (see Figure~\ref{fig:mem-example}). It confirms our intuition: the samples from the over-represented group fall into the low-memorization regime, atypical samples from the over-represented group (and a few similar samples from the under-represented group) compose the mid-level memorization regime, and the under-represented samples belong to the high-memorization regime. Finally, we observe that the correlation in this case is better; this suggests that memorization is a better indicator of MIA susceptibility.

\noindent{\bf Implications:} The aforementioned observations suggests that memorization values are a stronger indicator of MIA susceptibility. This suggests that propensity to be memorized is a better characteristic of the data that can explain its susceptibility to MIAs.

Prior work~\cite{carlini2022privacy} notes that outlier samples are more susceptible to MIAs. Through our analysis, we note that {\em not all outliers} are susceptible, and the current definition for "outliers" are unreliable. However, we observe that those outliers that are also highly likely to be memorized are definitively susceptible to MIAs. Thus, memorization provides a more nuanced understanding of the disparate MIA performance, and accurately characterizes MIA vulnerability.

%% file: sec_contents/mem_tab_figs.tex
\begin{figure*}[bht]
\centering
\begin{subfigure}{0.9\textwidth}
    \includegraphics[width=0.32\textwidth]{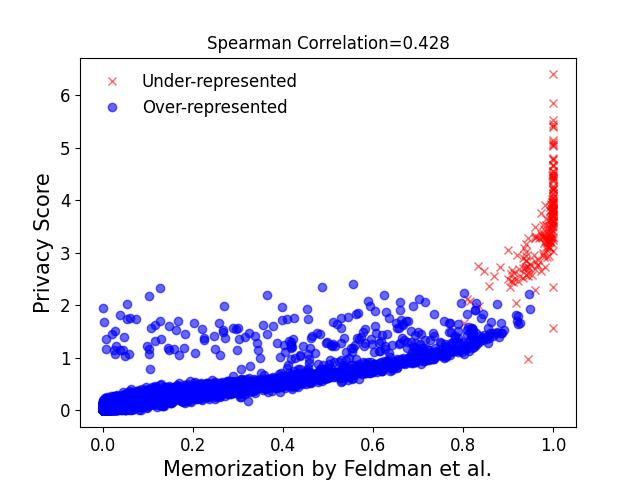}
    \includegraphics[width=0.32\textwidth]{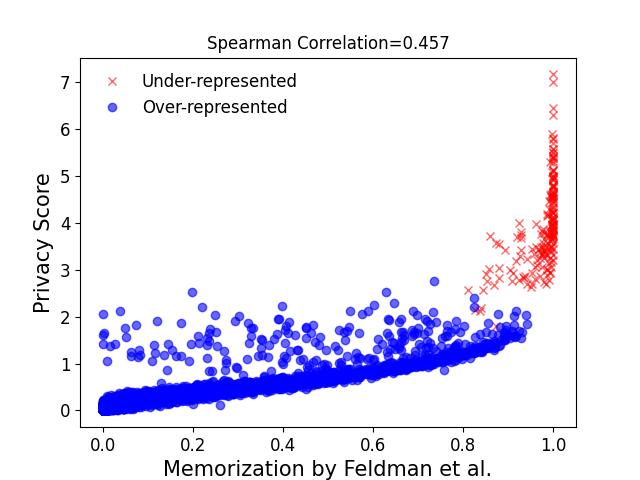}
    \includegraphics[width=0.32\textwidth]{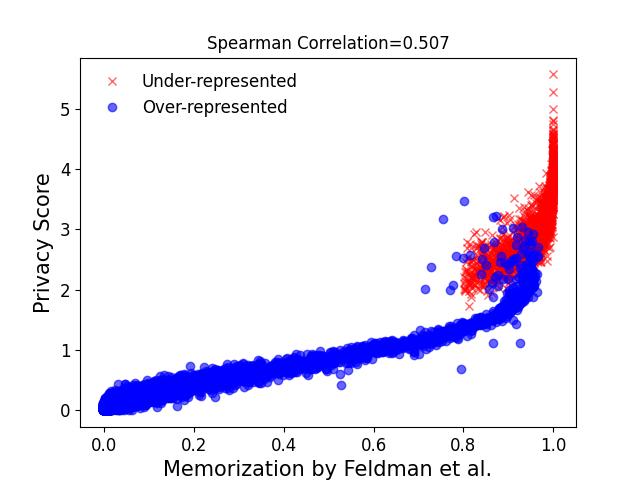}
    \caption{Privacy score vs. memorization}
    \label{fig:privacy-mem-singleton_top}
\end{subfigure}
\begin{subfigure}{0.9\textwidth}
    \includegraphics[width=0.32\textwidth]{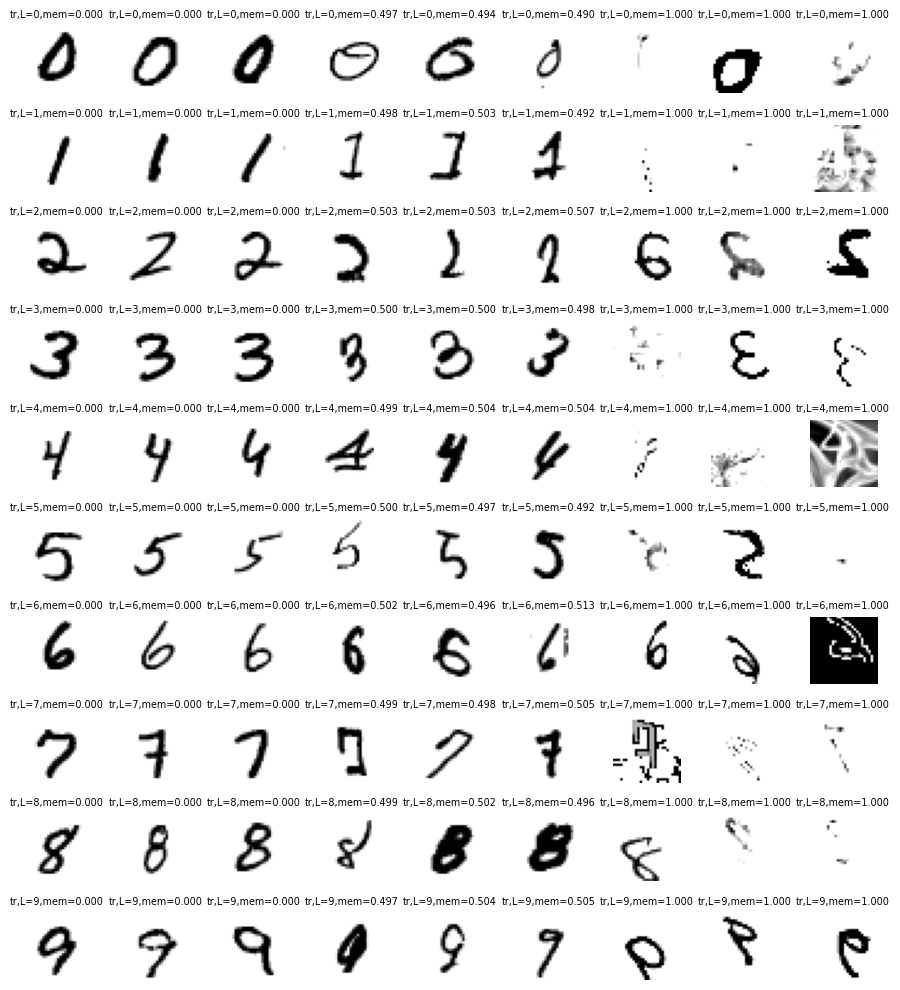}
    \includegraphics[width=0.32\textwidth]{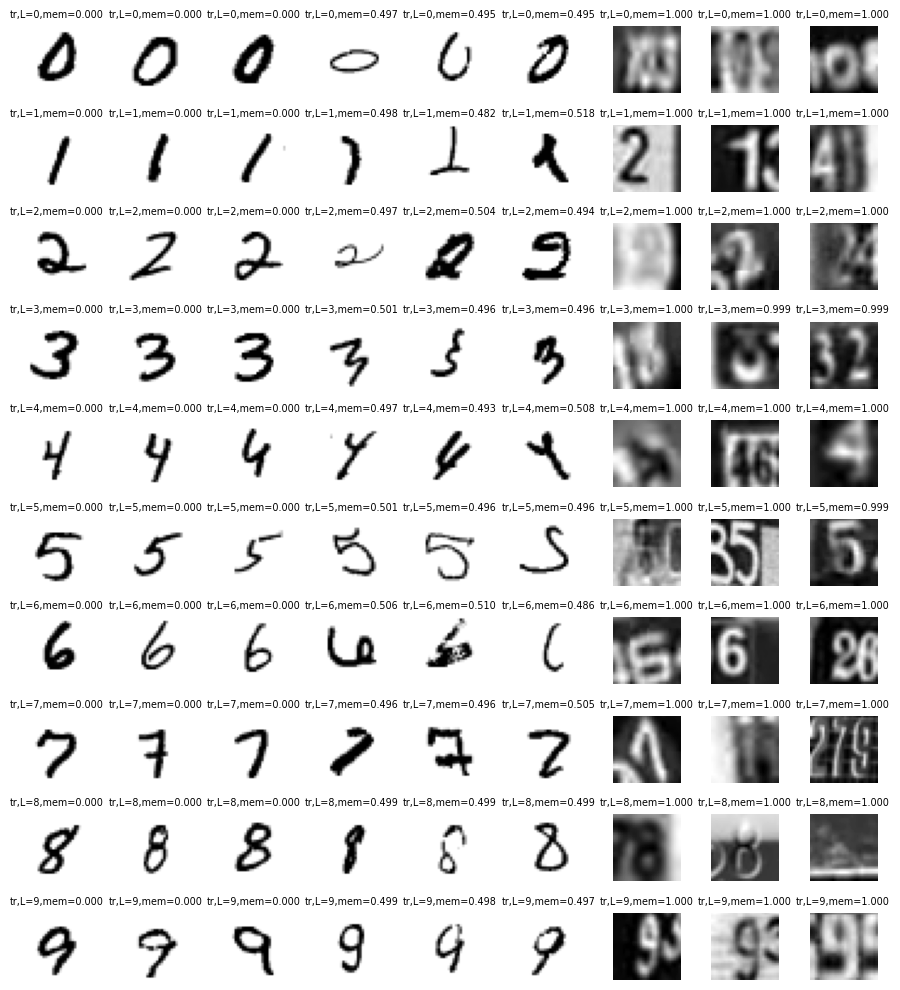}
    \includegraphics[width=0.32\textwidth]{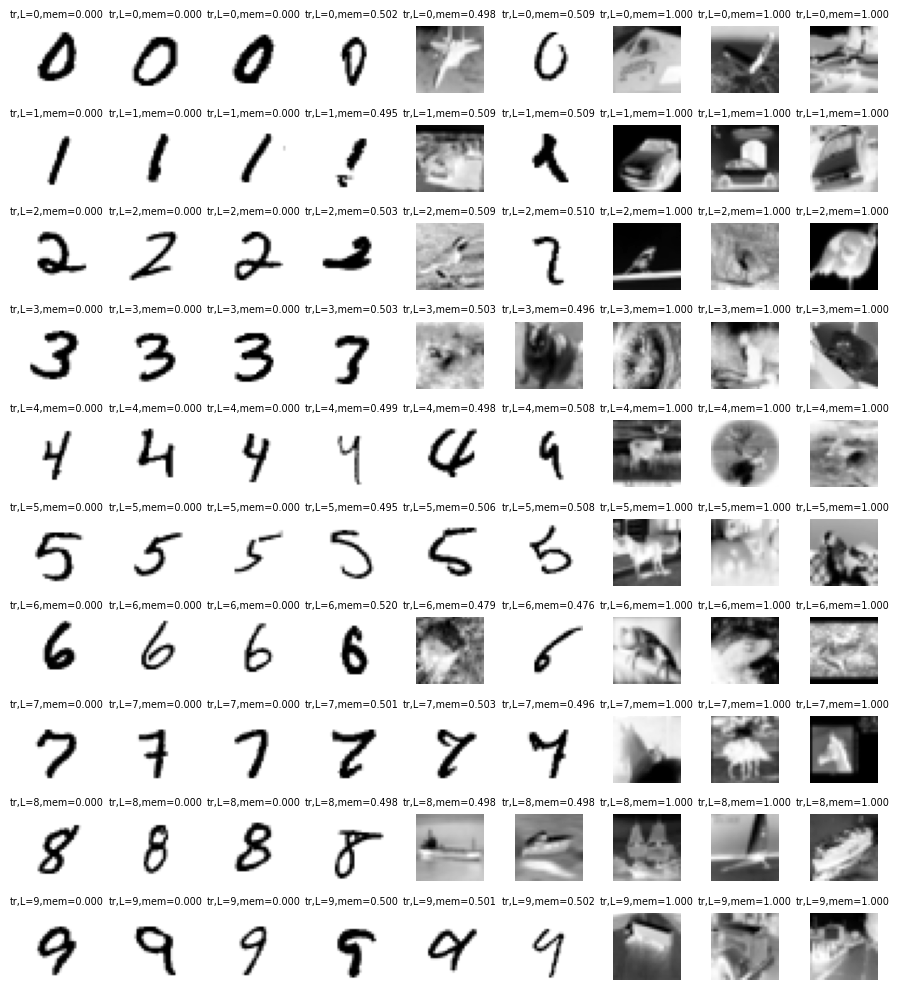}
    \caption{Random examples whose memorization value is close to 0 (first three columns), 0.5 (next three columns), 1 (last three columns). Each row shows images belonging to each of 10 classes.}
    \label{fig:mem-example}
\end{subfigure}
\caption{(i) \textbf{L}: MNIST+augmented MNIST ({\bf V1}), (ii) \textbf{M}: MNIST+SVHN ({\bf V2}), (iii) \textbf{R}: MNIST+CIFAR-10 ({\bf V3})}
\label{fig:mem-privacy-singleton}
\end{figure*}


%% file: sec_contents/5_theory_jihye.tex
\section{Membership Inference and Memorization}
\label{sec:theory}

In light of our preliminary findings on the interplay between label memorization and MI susceptibility in \S~\ref{sec:motivation}, we aim to establish a formal understanding of this connection.  We present a theoretical analysis that addresses two key aspects: (a) the memorization score of a point serves as a lower bound for the MI adversary's success (in determining membership of that point) under certain assumptions, and (b) higher values of memorization score correlate with improved computational efficiency of MIAs. To achieve these conclusions, we view the MI game through the lens of hypothesis testing. Our work aligns with recent endeavors to connect MIAs and hypothesis testing~\cite{ye2021enhanced, carlini2022membership}, and we provide a rigorous description of this connection, exploring various assumptions regarding the adversary's capabilities.

\noindent{\bf Note:} Recall that there are two variations in the design of the MI game as introduced in \S~\ref{sec:background}: (a) the challenger picks the dataset $S$ \ie corresponding to game $MI_{L,\calA}$~\cite{yeom2018privacy}, and (b) the adversary $\calA$ controls $S$ \ie corresponding to $MI^s_{L,\calA}$~\cite{mahloujifar2022optimal}.

\subsection{$MI_{L,\calA}$: Challenger Controls the Dataset}
\label{sec:game1}

Let $f: \Theta \rightarrow \mathbb{R}^\ell$ (\eg $f$ might extract the logits corresponding to the pre-softmax layer of a network~\cite{carlini2022membership}). The adversary $\mathcal{A}$ lacks direct access to the challenger-collected dataset $S$. However, leveraging the learning algorithm $L$, the adversary can initiate a random process involving sampling as follows: sample a new dataset $\hat{S} \sim \calD$ and can either, 
\begin{enumerate}
\item Execute $L$ on $\hat{S}$ to obtain $\theta_L \in \Theta$. Then output $f(\theta_L)$, which is characterized by distribution $P_0$. Note that $P_0$ on $\mathbb{R}^\ell$ depends on $\calD$ and also the randomness of $L$.
\item Execute $L$ on $\hat{S} \cup \{ z \}$ to obtain $\theta_L \in \Theta$. Then output $f(\theta_L)$, which is characterized by distribution $P_1$.
\end{enumerate}

\noindent{\bf Problem to solve:} $\calA$ receives $\theta_L \in \Theta$ and needs to decide if $f(\theta_L) \in \mathbb{R}^\ell$ was generated by distribution $P_0$ or $P_1$. In other words, $\calA$ needs a hypothesis test $T_{\calD, L}$ that, given \textit{one} sample $o \in \mathbb{R}^\ell$, outputs $0$ or $1$ corresponding to distributions $P_0$ and $P_1$. We define the following terms:
\begin{align*}
    \mathcal{T}_1 &= \Pr_{o \sim P_0} (T_{\calD,L} \; = \; 0) \\
    \mathcal{T}_2 &= \Pr_{o \sim P_1 } (T_{\calD,L} \; = \; 0)
\end{align*}
\ie the adversary in this MI game launches the hypothesis testing with $T_{\calD, L}$, and 
this gives us the following bound: $Adv^{T} (L,\calA) =  \mathcal{T}_1 - \mathcal{T}_2$
where $Adv^{T}(L,\calA)$ is $Adv_1(T_{\calD,L})$ (refer Equation~\ref{eq:adv_HT}). Next, we outline the specific hypothesis tests that the adversary can employ to launch effective MIAs in two scenarios.

\mypara{Scenario 1: $\calA$ perfectly knows $P_0$ and $P_1$.}
Assume that the adversary knows the density function for $P_0$ and $P_1$. 
Given $\theta_L$, $\calA$ can use the $\text{sLLR}$ test to output $0$ and $1$. Then we immediately get the following theorem by Lemma~\ref{lem:sLLR}.
\begin{tcolorbox}
\begin{thm}
\label{thm:sLLR-adv}
    $\mbox{Adv}  (L,\calA) ~\geq~ H^2 (P_0, P_1) $ \\
    where $\calA$ uses $\text{sLLR}$ test as described above. 
\end{thm}
\end{tcolorbox}

\mypara{Scenario 2: $\calA$ does not know $P_0$ and $P_1$.} Consider the case when the adversary only knows $P_0$ and $P_1$ implicitly, and does not know their probability density functions.
Then $\calA$ can make a parametric assumption on $P_0$ and $P_1$; for instance, assume that they are two $l$-dimensional multivariate normal distributions $N(\mu_1,\Gamma_1)$ and $N(\mu_2,\Gamma_2)$~\footnote{Such parametric approximation has been used in the literature~\cite{carlini2022membership}. Note that other than multivariate normal distribution, any parametric distribution (\eg exponential) can be assumed for
$P_0$ and $P_1$. The only requirement would be that the parameters of the distribution should be efficiently estimated, and preferably, the Hellinger distance should be expressed in a concise, closed form.}.
$\calA$ decides the output based on the following strategy:
\begin{enumerate}
    \itemsep0em
    \item Empirically estimate the parameters of $P_0$ and $P_1$. Generate $m$ datasets $\hat{S}_1,\cdots,\hat{S}_m$ sampled from $\calD$ (recall that $\calA$ knows $\calD$). Execute $L$ on $\hat{S}_1,\cdots,\hat{S}_m$  and $\hat{S}_1 \cup \{ z \},\cdots, \hat{S}_m \cup \{ z \}$ and generate two sets:
    $Z_0 = \{ f(\theta_1), \cdots, f(\theta_m) \}$ and $Z_1 = \{ f(\theta_{1,z}), \cdots, f(\theta_{m,z}) \}$. Then, estimate the corresponding parameters \ie ($\mu_0$, $\Gamma_0$) for $Z_0$. and ($\mu_1$, $\Gamma_1)$ for set $Z_1$. Note that $m$ corresponds to the number of shadow models in the literature~\cite{shokri2017membership, sablayrolles2019white, carlini2022membership}. $Z_0$ represents the \textbf{out} case where the data $z$ is excluded during training, and $Z_1$ represents the 
    \textbf{in} case where $z$ is a member of the training dataset. 
    \item Decide whether $f(\theta_L)$ given by the challenger is in $P_0$ and $P_1$ using $\text{sLLR}(f(\theta_L))$ test with normal distributions parameterized by $N(\mu_0,\Gamma_0)$ and $N(\mu_1,\Gamma_1)$. 
\end{enumerate}
The Hellinger distance $H^2 (P_0, P_1)$ where $P_0 \sim N(\mu_0,\Gamma_0)$ and $P_1 \sim N(\mu_1,\Gamma_1)$ is given by:
\begin{align*}
    1 ~-~ &\frac{ (\det(\Gamma_0) \det(\Gamma_1))^{\frac{1}{4}}}{\det(\frac{\Gamma_0+\Gamma_1}{2})^\frac{1}{2} }  \times \zeta    
\end{align*}
where $\zeta=\exp\left\{  -\frac{1}{8} (\mu_0-\mu_1)^{\top} ( \frac{\Gamma_0+\Gamma_1}{2} )^{-1} (\mu_0-\mu_1) \right\}$ (where $\top$ denotes transpose). This immediately becomes the lower bound for MI advantage by Theorem~\ref{thm:sLLR-adv}.

\noindent{\bf Note:} Recall that there is a one-sided version of the problem as well: $\calA$ receives $\theta_L \in \Theta$ and needs to decide if $f(\theta_L)$ was generated by distribution $P_0$ or not. In other words, the second distribution $P_1$ is absent.

\subsection{$MI^s_{L,\calA}$: Adversary Controls the Dataset}
\label{sec:adv-mem}

Assume that $\mathcal{A}$ has access to not only data distribution $\calD$ and learning algorithm $L$ but also to the dataset $S$. Symmetrically, the MI advantage in Equation~\ref{eq:adv_MI} can be rewritten as 
\begin{align}
\label{eq:adv_strong}
\mbox{Adv}  (L,\calA) ~&=~ \Pr(b_{\calA} = 1 \mid b_{\mathcal{C}} = 1) - \Pr(b_{\calA}=1 \mid b_{\mathcal{C}}=0) \nonumber\\
&\geq~ \Pr(\calA (L(S \cup \{ z \})) = 1) - \Pr(\calA (L(S)) = 1)
\end{align}
where the probability is over the randomness in $L$ and $\calA$.

In the above expression, $\calA$ is the shorthand for the algorithm used by the adversary to predict the bit $b_A$.
It can be viewed as a randomized function from $\Theta \times \calX$ to $\{ 0,1 \}$. 
Let $r_L$ and $r_{\calA}$ be the random strings corresponding to $L$ and $\calA$. 
If $r_L$ and $r_{\calA}$ are independent, we can rewrite Equation~\ref{eq:adv_strong} as
\begin{align}
\label{eq:adv_strong_2}
\Pr_{\theta \in L(S \cup  \{ z \})} \; \alpha(\calA,\theta,z) - \Pr_{\theta \in L(S) } \; \alpha(\calA,\theta,z)
\end{align}
where $\alpha(\calA,\theta) = \Pr_{r_{\calA}} (\calA(\theta,z) = 1)$. Note that $\alpha(\calA,\theta,z)$ is a \textit{deterministic} function from $\Theta \times \calX$ to $\{0, 1 \}$. This is a standard argument in analyzing games in cryptography; w.l.o.g, we can consider deterministic adversaries. 

Now, let us consider the following adversary: $\calA(\theta,z) = \mathbb{1}_{\theta(x) = y}$, where $z=(x,y)$; $x$ is data and $y$ is the corresponding label. 
In this case, Equation~\ref{eq:adv_strong_2} simplifies to:
\begin{align*}
    Adv^{m} (L,\calA) ~&=~ \Pr_{\theta \in L(S \cup  \{ z\})} (\theta(x) = y) ~-~ \Pr_{\theta \in L(S) }(\theta(x) = y) \\
    ~&= \mem(L,S,z) 
\end{align*}
which immediately proves that:
\begin{tcolorbox}
\begin{thm}
\label{thm:adv_mem}
$\mbox{Adv}  (L,\calA) ~\geq~ \mem(L,S,z)$. \\
This implies that if the adversary can choose $z$ with a high memorization score~\footnote{Caveat: in practical setup, we assume the existence of memorization oracle $\mathcal{O}_{\text{mem}}$ that gives the approximate memorization score given a sample. We leave the realization of a tractable, and effective $\mathcal{O}_{\text{mem}}$ as an important future work.}, the MI advantage can be increased.
\end{thm}
\end{tcolorbox}
\noindent Recall that for the advantage $\mbox{Adv}  (L,\calA)$ we implicitly take the supremum over adversaries in a certain class. Hence, an advantage $\beta$ with a specific adversary becomes a lower bound on the advantage of the adversary (see Appendix~\ref{app:theory} for advantage with other various instantiations of the adversary).

\subsection{Memorization and MIA Efficiency}
\label{sec:sample-mem}

Finally, based on our analysis in the above two subsections, we provide the connection between the memorization and the sample efficiency of MI game. 

Observe that Scenario 2 in \S~\ref{sec:game1} is the one which is most general \ie the adversary knows nothing about both distributions. In such settings, note that the adversaries need to train $m$ shadow models to obtain observations (or "samples") which can be used to approximate distributions, and then hypothesis testing is performed using these approximated distributions. Recall that Lemma~\ref{lem:opt_sample} (refer \S~\ref{subsec:hypo_testing}) defines the optimal sample complexity for any hypothesis test based on the Hellinger distance between two distributions. {\em Thus, one can observe that the number of shadow models corresponds to the sample complexity of the hypothesis test.} Two natural questions emerge: (a) how does one estimate this sample complexity when the distributions are unknown?, and (b) how many samples do we need to effectively estimate the distribution for hypothesis testing?\footnote{Both questions have the same implication.} To answer the questions, we utilize an observation made by Feldman~\cite{feldman2020longtail}: the definition of memorization {\em is exactly the} TV distance between the distributions of the indicator of the label being $y$ (or not). Based on the earlier inequality connecting TV and Hellinger distances, we know that
\begin{equation}
\label{eq:hellinger_mem}
H^2(P_0,P_1) \leq \mem(L,S,z)
\end{equation}

By plugging Equation~\ref{eq:hellinger_mem} into Lemma~\ref{lem:opt_sample}, we get
\begin{tcolorbox}
\begin{corollary}
\label{corl:sample_mem}
$M_\alpha^{P_0, P1} (T^\star) ~=~ \Tilde{\Theta} \left( \frac{1}{\mem(L,S,z)} \right)$.
\end{corollary}
\end{tcolorbox}
where $\Tilde{\Theta}$ captures constants associated with the inequality. The implication is that if $\calA$ can launch the attack using $z$ with high memorization score, the required number of shadow models can be reduced, which is the major computational bottleneck in most popular attacks~\cite{song2021systematic, sablayrolles2019white, shokri2017membership, carlini2022membership}.

%% file: sec_contents/6_implementation.tex
\section{Implementation Details}
\label{sec:implementation}

Having established the theoretical connection between memorization, MIAs and hypothesis testing, and showing how it can help reduce training run-time, we briefly describe our experimental setup used to validate the theory. More details can be found in the Appendix~\ref{app:implementation-details}.

\noindent{\bf Datasets.}
Similar to the setup described in \S~\ref{sec:motivation}, we consider a dataset $\Dtr$ that comprises a mixture of two distinct groups. Throughout our experiments, we set $\Dotr$ as either the MNIST train dataset (consisting of 60,000 samples)~\cite{lecun2010mnist} or the CIFAR-10 train dataset (consisting of 50,000 samples)~\cite{cifar10}. We explore three different approaches to define $\Dutr$: selecting (a) random OOD samples, (b) random OOD samples that belong to a particular subpopulation, and (c) highly memorized OOD samples. Particularly:
\begin{enumerate}[label=(\alph*)]
\itemsep0em
\item ({\bf Random}) When $\Dotr$ is the MNIST train set, $\Dutr$ is 1000 MNIST samples augmented by approaches proposed by Hendrycks \etal~\cite{hendrycks2022pixmix}, 1000 SVHN samples, or 6000 CIFAR-10 samples. When $\Dotr$ is CIFAR-10, then $\Dutr$ is 1000 samples randomly drawn from the CIFAR-100 train set~\cite{cifar10}, or 1000 random CIFAR-10 train data augmented as earlier~\cite{hendrycks2022pixmix}. Labeling across both groups was consistent. We ensure that the accuracy (both test and train) on both the under and over-represented groups is reasonable.
\item  ({\bf Subpopulations}) To identify subpopulations within the under-represented groups, we adopt the methodology proposed by Jagielski \etal~\cite{jagielski2021subpopulation}. For each construction of $\Dtr$ as specified in (a), we first preprocess $\Dtr$ by extracting the feature representations from the penultimate layer of a DNN trained on $\Dtr$ (\eg Efficientnet~\cite{tan2019efficientnet} in our case). Then, we apply a PCA projection to further reduce the feature dimensionality to 10. Finally, the KMeans clustering algorithm~\cite{kmeans} is applied to cluster the projected $\Dtr$ into five subgroups. We identify those samples from $\Dutr$ belonging to the smallest cluster as the desired subpopulation~\footnote{We note that hyperparameter choices, such as dimensionality of projection space, number of clusters, etc., are left to the adversary's specifications depending on the setup, and one could always come up with other methodologies to instantiate the subgroup identification.}.
\item ({\bf Singletons}) To instantiate the memorization oracle $\mathcal{O}_{mem}$, we use the approximation algorithm proposed by Feldman and Zhang~\cite{feldman2020neural}. For each choice of $\Dotr \bigcup \Dutr$ as specified in (a), we define {\em singletons} as data belonging to $\Dutr$ whose memorization degree exceeds a threshold ($0.8$ when $\Dotr$ is MNIST, and $0.5$ when $\Dotr$ is CIFAR-10~\footnote{To ensure an adequate representation of singletons in conjunction with the over-represented population, we choose the threshold with which $|\Dutr|$ is not significantly smaller than $|\Dotr|$}).
\end{enumerate}

We choose these 3 categories of obtaining subgroups as they represent popular theories for explaining the disparate performance of MIAs on samples. Evaluation on these categories will help better understand which theory is more accurate in explaining the behavior. 
Note that while these datasets may seem artificially mixed, our comparison is not the ``absolute'' success of any single attack, but relative gains for each attack between different types of vulnerable samples (to truly understand the best underlying cause).

\noindent{\bf Model Architectures.} 
We employ classical architectures commonly used in the computer vision community: two CNN models with varying convolutional filter sizes set to 32 and 64 (as in Carlini \etal~\cite{carlini2022membership}),
and two ResNet models (ResNet-9 and ResNet-18)~\cite{resnet}. Our primary focus is on the vision setting, as the notion of label memorization is well-studied and evaluated in this setting.

\noindent{\bf MIAs.}
We consider five representative attacks from the MIA literature~\cite{yeom2018privacy, shokri2017membership, song2021systematic, sablayrolles2019white, carlini2022membership}. 
Yeom \etal~\cite{yeom2018privacy} employ a very simple attack by thresholding the loss output without necessitating the adversary's access to the training dataset or distribution. On the other hand, other MIAs are built upon the assumption that the MI adversary has query access to the training data distribution for shadow model training~\cite{shokri2017membership,song2021systematic,sablayrolles2019white,carlini2022membership}.
For those attacks, we train 2000 shadow models for each setting.
Throughout our evaluation, we consider the strongest adversary who has control over the construction of the training data distribution (refer \S~\ref{subsec:ml}).

\noindent{\bf Metrics.}
Based on the premise established by Carlini \etal~\cite{carlini2022membership}, for each attack, we report the TPR when a decision threshold $\tau$ is chosen to maximize TPR at a target FPR (\eg as low as $0.1\%$ in our setting) along with AUROC values (\ie the probability that a positive example would have higher value than a negative one).

\noindent{\bf Disclaimer:}
Our evaluation setup spans 7 datasets and 4 models, and 5 MIAs. The magnitude of this evaluation is comparable to Carlini \etal~\cite{carlini2022membership}. We stress that performing such rigorous evaluation is also computationally expensive, requiring the training of a cumulative total of 38000 shadow models (2000 shadow models for each dataset and each variation of under-represented data selection). 

%% file: sec_contents/7_results.tex
\section{Empirical Validation}
\label{sec:results}

Having established the theoretical connections, we now validate them. Our experiments are designed to answer the following questions:
\begin{enumerate}
\itemsep0em
\item Are MIAs more performant on samples with high memorization scores compared to randomly chosen OOD samples or samples that belong to under-represented subpopulations?
\item If the adversary knows that a particular sample is likely to have a high memorization score, can they launch more computationally efficient MIAs?
\end{enumerate}
Based on our experiments, we observe that:
\begin{enumerate}
\itemsep0em
\item MIAs that utilize highly memorized data are significantly more effective than those based on other phenomenon \ie memorization can accurately explain the disparate performance of MIAs; the attack of Carlini \etal~\cite{carlini2022membership} with highly memorized data consistently achieves an AUROC of 1.0 across datasets. Moreover, it consistently exceeds TPR of $96\%$ (at $0.1\%$ FPR) and, in some cases, even attains a perfect TPR of $100\%$ (Table~\ref{tab:mia-results-mnist} and Table~\ref{tab:mia-results-cifar10}). Details are presented in \S~\ref{subsec:success}.
\item From Figure \ref{fig:num_shadow_mnist_mem_under} and Figure~\ref{fig:num_shadow_cifar10_mem_under}, we can see that an adversary that utilizes data with high memorization scores can achieve a reduction in MIA overhead (\ie training fewer shadow models). Specifically, across all considered dataset choices, the adversary achieves an equivalent level of attack performance with mere 5 shadow models to match the performance with 2000 shadow models under the same setup, a 400$\times$ reduction. The same effect is observed even when the adversary utilizes random OOD data (Figure \ref{fig:num_shadow_mnist_ood_under} and Figure~\ref{fig:num_shadow_cifar10_ood_under}), albeit not the same magnitude of reduction. Details are presented in \S~\ref{subsec:overheads}.
\end{enumerate}
We assume that the adversary has access to an oracle $\mathcal{O}_\text{mem}$ which provides memorization scores for any sample $z$. In the following sections, we explain the aforementioned answers.

\input{sec_contents/results_tables/table2}

\subsection{Increased MIA Success}
\label{subsec:success}



Table~\ref{tab:mia-results-mnist} and Table~\ref{tab:mia-results-cifar10} provide a comprehensive summary of the results obtained from evaluating MIAs across various data settings (see Table~\ref{tab:mia-results-mnist-additional} in Appendix~\ref{app:additional-results-mia} for more results).

For every attack, the first row corresponds to the performance of the attacks in the conventional setting, where $\Dtr$ is constructed using a single dataset. Subsequently, the next group of three rows represent scenarios where the adversary constructs the training dataset as a mixture of two groups, with variations in the selection of the under-represented group, as outlined in \S~\ref{sec:implementation}. For each setting, we provide the attack success rate under two conditions: (a) when random data is used as the challenge data (denoted as column ``All"), as commonly used in literature, and (b) when the adversary specifically utilizes data from the under-represented group (denoted as column ``Under-represented")~\footnote{When dataset consists of single population (\eg MNIST, or CIFAR-10), ``Under-represented" reports the attack performance on the data samples with memorization scores above 0.4 for MNIST, and 0.8 for CIFAR-10; the smaller threshold is chosen for MNIST since the dataset is less long-tailed. This is to observe if the adversary can still leverage memorization to launch a better attack even in a single dataset setup.}. This comparative analysis allows for a deeper understanding of MIA effectiveness by examining the impact of using data from the under-represented group. 

\noindent{\bf Takeaway 1:} We observe that every attack {\em always performs better} with respect to the sample from the under-represented group (\ie OOD samples, OOD subpopulation, or OOD singletons) than with a random sample (most likely to be drawn from the over-represented groups); given MNIST+augMNIST (random), the attack proposed by Shokri \etal~\cite{shokri2017membership} yields an overall AUROC of $0.51$ and TPR of $0.20\%$ under ``All''. However, we see an increase in the AUROC to $0.72$ and TPR to $0.70\%$ when the attack is evaluated on under-represented augMNIST samples (under ``Under-represented'' of MNIST+augMNIST (random) setting). {\em The most substantial improvement is observed when utilizing OOD singletons in all scenarios}; in the case of MNIST+SVHN (singletons), the attack devised by Sablayrolles \etal~\cite{sablayrolles2019white} demonstrates enhanced efficacy from an AUROC of 0.51 and TPR of 2.90\% (see ``All'' of MNIST+SVHN (singletons)) to an AUROC of 1.0 and TPR of 97.55\% (see ``Under-represented'' of MNIST+SVHN (singletons)). This surpasses the attack's performance when evaluated using OOD samples, where the AUROC is 0.81 and TPR is 2.86\% (see ``Under-represented'' of MNIST+SVHN (random)), or even using OOD subpopulation, where the AUROC is 0.95 and TPR is 26.0\% (see ``Under-represented'' of MNIST+SVHN (subpopulation)).

While we see increased efficiency with singletons in most of the cases, we note that utilizing samples from OOD subpopulations leads to better attack success in some cases. Specifically, see the results of Yeom \etal~\cite{yeom2018privacy} with CIFAR-10+CIFAR-100 dataset in Table~\ref{tab:mia-results-cifar10}. One possible explanation for this phenomena could be made by observing the distribution of memorization scores of samples in CIFAR-10+CIFAR-100 dataset (see Figure~\ref{fig:varun} in Appendix~\ref{app:additional-results-mia}). We observe that the distribution of memorization scores for the original CIFAR-10 dataset is long-tailed (as in Figure~\ref{fig:mem-distr-cifar10}), but when mixed with a random subset of CIFAR-100 dataset, all samples fall into the lower memorization regime under the score of $0.6$ (as in Figure~\ref{fig:mem-distr-cifar10+cifar100}). That is, even if the adversary constructs the under-represented group using the threshold $0.5$, the identified samples are not really highly memorized, and eventually, not any better than being selected by subpopulation identification. 

\noindent{\bf Takeaway 2:} Nonetheless, it is worth noting that in all scenarios, the attack proposed by Carlini \etal~\cite{carlini2022membership} consistently outperforms other attacks. This observation is consistent with the findings reported in their original paper.
Moreover, their attack achieves the best efficacy with singletons, even when singleton identification cannot be made reliably (as in the case of CIFAR10 + CIFAR100 dataset). 
Notably, by utilizing singletons, the attack achieves near-perfect AUROC and TPR even at a low FPR of $0.1\%$. We believe that the remarkable performance of this attack can be attributed to the inherent connection between the attack itself and the estimation of memorization (further discussion can be found in \S~\ref{sec:discussion}).

\begin{tcolorbox}
{\bf Result:} With the knowledge of memorization scores, the performance of MIAs improves significantly.
\end{tcolorbox}

\input{sec_contents/results_tables/shadow_models}

\subsection{Decreased Computational Overhead}
\label{subsec:overheads}

In \S~\ref{sec:theory}, we theoretically demonstrated that for a sample with a large memorization score, the number of shadow models required to distinguish the \textbf{in} case (\ie when the sample is a member) from the \textbf{out} case (\ie when the sample is not a member) is low. In this section, we provide empirical evidence that this relation holds in practical settings.

In Figure~\ref{fig:num_shadow_singleton}, we plot the success of the MIA by Carlini \etal~\cite{carlini2022membership} against the number of shadow models needed to achieve it. We use this attack as it demonstrates the most success (refer to results in \S~\ref{subsec:success}). 
Specifically, Figure~\ref{fig:num_shadow_mnist_ood_under} and Figure~\ref{fig:num_shadow_cifar10_ood_under} represent scenarios where the model is trained partly using OOD data \ie $D_U^{tr}$ comprises of random OOD data. Similarly, Figure~\ref{fig:num_shadow_mnist_mem_under} and Figure~\ref{fig:num_shadow_cifar10_mem_under} illustrate cases where the model is trained partly using data from highly memorized OOD data. 
For the comparison with the cases where the attack is employed using data randomly chosen without consideration of memorization, see Figure~\ref{fig:num_shadow_overrepresented} in Appendix~\ref{app:additional-results-mia}.

\noindent{\bf Takeaway 1:} Observe that for high memorization score samples, there is a substantial reduction in the number of shadow models needed to achieve a particular success threshold. For example, in Figure~\ref{fig:num_shadow_cifar10_mem_under}, it is phenomenal that we could achieve AUROC of $1.0$ with just one shadow model for \textbf{in} case and \textbf{out} case each, without necessitating thousands of shadow models. 

\noindent{\bf Takeaway 2:} We also see that the same phenomenon holds even when the adversary leverages random OOD data. We observe that less than 50 shadow models are sufficient to achieve the same level of AUROC as 2000 shadow models (see Figure~\ref{fig:num_shadow_mnist_ood_under} and Figure~\ref{fig:num_shadow_cifar10_ood_under}), a 40$\times$ reduction. 
These findings suggest that even in situations where the adversary does not have access to a reliable memorization oracle, there is still potential to reduce the computational overhead of MIAs, although to a lesser extent than when utilizing highly memorized samples. We acknowledge that this observation is purely empirical, and does not have any theoretical backing.

\begin{tcolorbox}
{\bf Result:} With the knowledge of memorization, the required computational overhead for training shadow models reduces significantly.
\end{tcolorbox}

%% file: sec_contents/results_tables/table2.tex
\begin{table*}[h]
\begin{adjustbox}{width=0.9\textwidth,center}
\begin{tabular}{l|l|cc|cc}
\toprule
\multirow{2}{*}{\bf Method} & \multirow{2}{*}{{\bf Dataset} ($\Dtr$)} & \multicolumn{2}{c|}{{\bf AUROC} $\uparrow$} & \multicolumn{2}{c}{{\bf TPR @ $0.1 \%$ FPR} $\uparrow$} \\ \cline{3-6}
                                & \multicolumn{1}{c|}{} & \multicolumn{1}{c}{All} & Under-represented    & \multicolumn{1}{c}{All} & Under-represented \\ 
\hline \hline
\multirow{7}{*}{Yeom \etal~\cite{yeom2018privacy}} & MNIST & 0.50 & 0.50 & 0.0 \% & 0.0 \% \\ \cdashline{2-6}
                                & MNIST+augMNIST (random) & 0.50  & 0.61 & 0.0 \% & 0.0 \%  \\
                                & MNIST+augMNIST (subpopulation) & 0.50 & 0.80 & 0.0 \% & 0.0 \%\\
                                & MNIST+augMNIST (singletons) & 0.50 & \textbf{0.82} & 0.0 \% & 0.0 \% \\ \cdashline{2-6}
                                & MNIST+SVHN  (random) & 0.50 & 0.77 & 0.0 \% & 0.0 \%  \\
                                & MNIST+SVHN  (subpopulation)& 0.50 & 0.80 & 0.0 \% & 0.0 \% \\
                                & MNIST+SVHN (singletons) & 0.50 & \textbf{0.93} & 0.0 \%                    & \textbf{2.86\%} \\
                                \cline{1-6}
                                
\multirow{7}{*}{Shokri \etal~\cite{shokri2017membership}} & MNIST & 0.50 & 0.66 & 0.07 \% & 3.33 \%\\ \cdashline{2-6}
                                & MNIST+augMNIST (random)& 0.51 & 0.72 & 0.20 \% & 0.70 \%  \\
                                & MNIST+augMNIST (subpopulation)& 0.51 & 0.66 & 0.12 \% & 3.25 \% \\
                                & MNIST+augMNIST (singletons) & 0.51  & \textbf{0.99} & 0.15 \% & \textbf{18.88 \%} \\ \cdashline{2-6}
                                & MNIST+SVHN (random)& 0.52 & 0.76 & 0.25 \% & 0.30 \%  \\
                                 & MNIST+SVHN (subpopulation) & 0.51 & 0.91 & 0.10 \% & 1.41 \% \\
                                & MNIST+SVHN (singletons) & 0.51 & \textbf{0.99} & 0.23 \% & \textbf{72.86 \%} \\
                                \cline{1-6}
\multirow{7}{*}{Sablayarolles \etal~\cite{sablayrolles2019white}} & MNIST & 0.51 & 0.75 & 0.12 \% & 15.09 \%\\ \cdashline{2-6}
                                & MNIST+augMNIST (random) & 0.51 & 0.85 & 0.22 \% & 0.61 \%  \\
                                & MNIST+augMNIST (subpopulation) & 0.52 & 0.93 & 0.11 \% & 27.91 \% \\
                                & MNIST+augMNIST (singletons) & 0.52 & \textbf{0.99} & 0.33 \% & \textbf{46.10 \%} \\ \cdashline{2-6}
                                & MNIST+SVHN (random)& 0.50 & 0.81 & 0.21 \% & 2.86 \%  \\
                                & MNIST+SVHN (subpopulation)& 0.51 & 0.95 & 0.14 \%& 26.0 \%\\
                                & MNIST+SVHN (singletons) & 0.51 & \textbf{1.0} & 2.90 \% & \textbf{97.55 \%} \\
                                \cline{1-6}

\multirow{7}{*}{Song \etal~\cite{song2021systematic}} & MNIST & 0.51 &0.68 & 0.14 \% & 2.96 \% \\ \cdashline{2-6}
                                & MNIST+augMNIST (random) & 0.51 & 0.70 & 0.16 \% & 0.53 \%  \\
                                & MNIST+augMNIST (subpopulation) & 0.50 & 0.92 & 0.11 \% & 11.57 \% \\
                                & MNIST+augMNIST (singletons) & 0.51 & \textbf{0.98} & 0.10 \% & \textbf{16.12 \%} \\ \cdashline{2-6}
                                & MNIST+SVHN (random)& 0.50 & 0.77 & 0.17 \% & 2.86 \%  \\
                                & MNIST+SVHN (subpopulation) & 0.51 & 0.91 & 0.14 \% & 9.18 \%\\
                                & MNIST+SVHN (singletons) & 0.51 & \textbf{0.98} & 2.04 \% & \textbf{14.11 \%} \\
                                \cline{1-6}                                
                                
\multirow{7}{*}{Carlini \etal~\cite{carlini2022membership}} & MNIST & 0.52 & 0.89 & 0.88\% & 45.98 \% \\ \cdashline{2-6}
                                & MNIST+augMNIST (random)& 0.53 & 0.87 & 1.45 \% & 27.55 \% \\
                                & MNIST+augMNIST (subpopulation)& 0.52 & 0.96 & 0.33 \% & 68.29 \% \\
                                & MNIST+augMNIST (singletons) & 0.53 & \textbf{1.0}  & 1.07 \% & \textbf{100 \%} \\ \cdashline{2-6}
                                & MNIST+SVHN (random)& 0.53 & 0.95 & 2.15 \% & 51.5 \% \\
                                & MNIST+SVHN (subpopulation)& 0.52 & 0.99 & 0.50 \% & 78.71 \%\\
                                & MNIST+SVHN (singletons) & 0.53 & \textbf{1.0}  & 1.24 \% & \textbf{100 \%} \\ \cline{1-6}
\end{tabular}%
\end{adjustbox}
\caption{\textbf{Comparison of representative MIAs.} 
For each choice of MIA and mixture dataset, we direct readers to compare AUROC and TPR between (i) (All) vs. (Under-represented), and (ii) (random) vs. (subpopulation) vs. (singletons) to see the effect of utilizing singletons in MIAs. Best results are \textbf{boldfaced}.}
\label{tab:mia-results-mnist}
\end{table*}

\begin{table*}[tbh]
\begin{adjustbox}{width=0.9\textwidth,center}
\begin{tabular}{l|l|cc|cc}
\toprule
\multirow{2}{*}{\bf Method} & \multirow{2}{*}{{\bf Dataset} ($\Dtr$)} & \multicolumn{2}{c|}{{\bf AUROC} $\uparrow$} & \multicolumn{2}{c}{{\bf TPR @ $0.1 \%$ FPR} $\uparrow$} \\ \cline{3-6}
                                & \multicolumn{1}{c|}{} & \multicolumn{1}{c}{All} & Under-represented    & \multicolumn{1}{c}{All} & Under-represented \\ 
\hline \hline
\multirow{6}{*}{Yeom \etal~\cite{yeom2018privacy} } & CIFAR-10 & 0.62
& 0.73 & 0.0 \% & 0.0 \% \\ \cdashline{2-6}
                                & CIFAR-10+augCIFAR-10 (random) & 0.57 & 0.69 & 0.0 \% & 0.0 \% \\
                                & CIFAR-10+augCIFAR-10 (subpopulation) & 0.57 & 0.69 & 0.0 \% & 0.0 \% \\
                                & CIFAR-10+augCIFAR-10 (singletons) & 0.57 & \bf 0.71 & 0.07 \% & \bf 0.59 \%\\ \cdashline{2-6}
                                & CIFAR-10+CIFAR-100 (random) & 0.57 & 0.79 & 0.0 \% & 0.0 \% \\
                                & CIFAR-10+CIFAR-100 (subpopulation) & 0.62 & \bf \textcolor{black}{0.85} & 0.0 \% & \bf \textcolor{black}{17.32} \% \\
                                & CIFAR-10+CIFAR-100 (singletons) & 0.63 & 0.76 & 0.0 \% & 0.0 \%\\
                                \cline{1-6}
                                
\multirow{6}{*}{Shokri \etal~\cite{shokri2017membership}} & CIFAR-10 & 0.69 & 0.99 & 0.41 \% & 53.70 \% \\ \cdashline{2-6}
                                & CIFAR-10+augCIFAR-10 (random) & 0.63 & 0.78 & 0.16 \% & 1.17 \% \\
                                & CIFAR-10+augCIFAR-10 (subpopulation) & 0.60 & 0.78& 0.12 \% & 0.76 \% \\
                                & CIFAR-10+augCIFAR-10 (singletons) & 0.63 & \bf 0.97 & 0.20 \% & \bf 39.29 \% \\ \cdashline{2-6}
                                & CIFAR-10+CIFAR-100 (random) & 0.69 & 0.85 & 0.21 \% & 0.28 \% \\
                                & CIFAR-10+CIFAR-100 (subpopulation) & 0.68 & 0.95& 0.24 \% & 20.16 \%\\
                                & CIFAR-10+CIFAR-100 (singletons) & 0.70 & \bf 1.0 & 0.45 \% & \bf 100 \% \\
                                \cline{1-6}

\multirow{6}{*}{Sablayarolles \etal~\cite{sablayrolles2019white}} & CIFAR-10 & 0.71 & 1.0 & 6.18 \% & 68.17 \% \\ \cdashline{2-6}
                                & CIFAR-10+augCIFAR-10 (random) & 0.68 & 0.87 & 3.20 \% & 9.17 \% \\
                                & CIFAR-10+augCIFAR-10 (subpopulation) & 0.67 & 0.87 & 3.17 \% & 6.03 \%\\
                                & CIFAR-10+augCIFAR-10 (singletons) & 0.72 & \bf 1.0 & 3.14 \% & \bf 46.97 \% \\ \cdashline{2-6}
                                & CIFAR-10+CIFAR-100 (random) & 0.74 & 0.90 & 5.97 \% & 37.05 \% \\
                                & CIFAR-10+CIFAR-100 (subpopulation) & 0.75 & 1.0 & 5.01 \% & 56.90 \% \\
                                & CIFAR-10+CIFAR-100 (singletons) & 0.75 & \bf 1.0 & 6.21 \% & \bf 82.31 \%\\
                                \cline{1-6}
                                
\multirow{6}{*}{Song \etal~\cite{song2021systematic}} & CIFAR-10 & 0.62 & 0.95 & 3.18 \% & 19.52\% \\ \cdashline{2-6}
                                & CIFAR-10+augCIFAR-10 (random) & 0.55 & 0.65 & 1.18 \% & 5.4 \% \\
                                & CIFAR-10+augCIFAR-10 (subpopulation) & 0.54 & 0.62 & 1.01 \% & 3.2 \% \\
                                & CIFAR-10+augCIFAR-10 (singletons) & 0.55 & \bf 0.81 & 3.15 \% & \bf 18.20 \% \\ \cdashline{2-6}
                                & CIFAR-10+CIFAR-100 (random) & 0.60 & 0.89 & 3.01 \% & 11.25 \%\\
                                & CIFAR-10+CIFAR-100 (subpopulation) & 0.60 & 0.92 & 2.89 \% & 15.08 \% \\
                                & CIFAR-10+CIFAR-100 (singletons) & 0.61 & \bf 0.98 & 3.13 \% & \bf 21.11 \% \\
                                \cline{1-6}
                                
\multirow{6}{*}{Carlini \etal~\cite{carlini2022membership}} & CIFAR-10 & 0.87 & 1.0 & 15.25 \% & 95.47 \% \\ \cdashline{2-6}
                                & CIFAR-10+augCIFAR-10 (random) & 0.73 & 0.93 & 5.96 \% & 36.73 \% \\
                                & CIFAR-10+augCIFAR-10 (subpopulation) & 0.72 & 0.93 & 2.81 \% & 29.55 \% \\
                                & CIFAR-10+augCIFAR-10 (singletons) & 0.74 & \bf 1.0 & 6.91 \% & \bf 98.01 \%\\ \cdashline{2-6}
                                & CIFAR-10+CIFAR-100 (random) & 0.87 & 0.96 & 14.62 \% & 54.42 \%\\
                                & CIFAR-10+CIFAR-100 (subpopulation) & 0.85 & 0.99 & 12.21 \% & 89.76 \% \\
                                & CIFAR-10+CIFAR-100 (singletons) & 0.86 & \bf 1.0 & 15.79 \% & \bf 100.0 \%\\
                                \cline{1-6}
\end{tabular}%
\end{adjustbox}
\caption{\textbf{Comparison of representative MIAs.} For each attack, we evaluate the performance on three different mixtures of datasets: CIFAR-10 with (a) augmented CIFAR-10, and (b) CIFAR-100. For each choice of MIA and mixture dataset, we direct readers to compare AUROC and TPR between (a) (All) vs. (Under-represented), and (b) (random) vs. (singletons) to see the effect of utilizing singletons in MIAs. Best results are \textbf{boldfaced}.}
\label{tab:mia-results-cifar10}
\end{table*}

%% file: sec_contents/results_tables/shadow_models.tex
\begin{figure*}[tbp]
  \centering
  

  \begin{subfigure}{0.38\textwidth}
    \includegraphics[width=\linewidth]{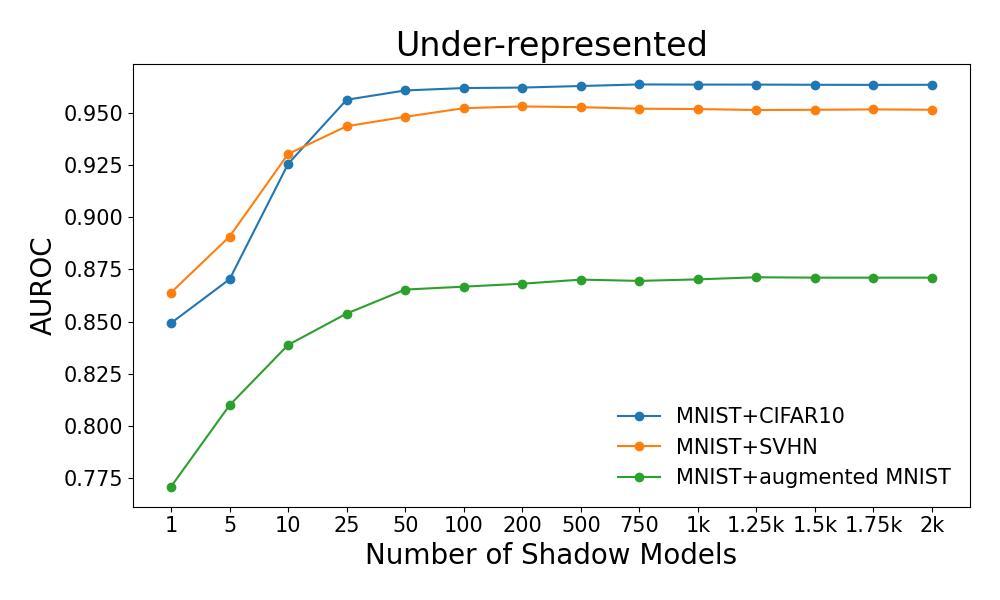}
    \caption{MNIST + OOD data}
    \label{fig:num_shadow_mnist_ood_under}
  \end{subfigure}
  \begin{subfigure}{0.38\textwidth}
    \includegraphics[width=\linewidth]{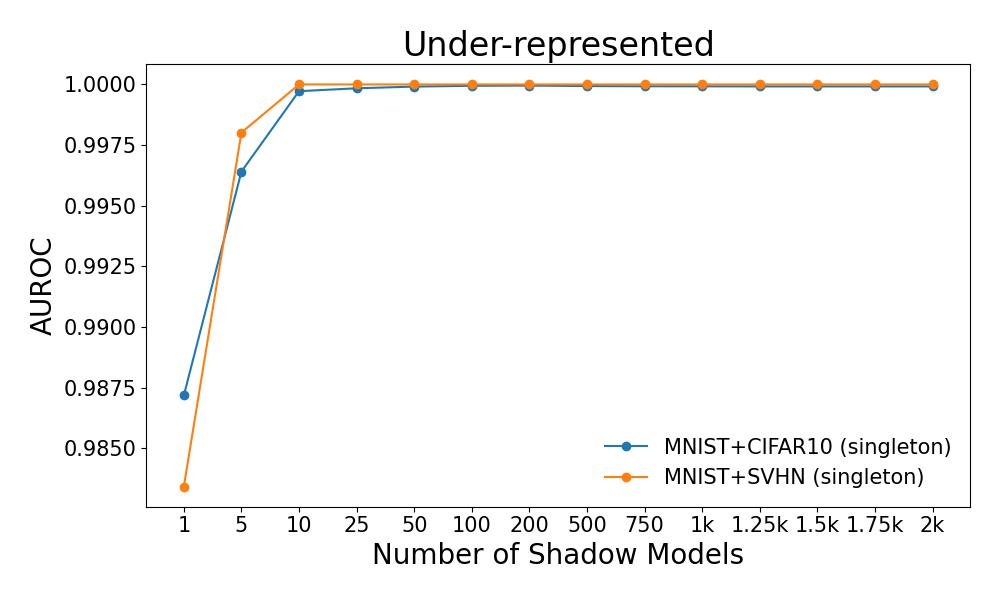}
    \caption{MNIST + HM OOD data}
    \label{fig:num_shadow_mnist_mem_under}
  \end{subfigure} 
  \\
  \begin{subfigure}{0.38\textwidth}
    \includegraphics[width=\linewidth]{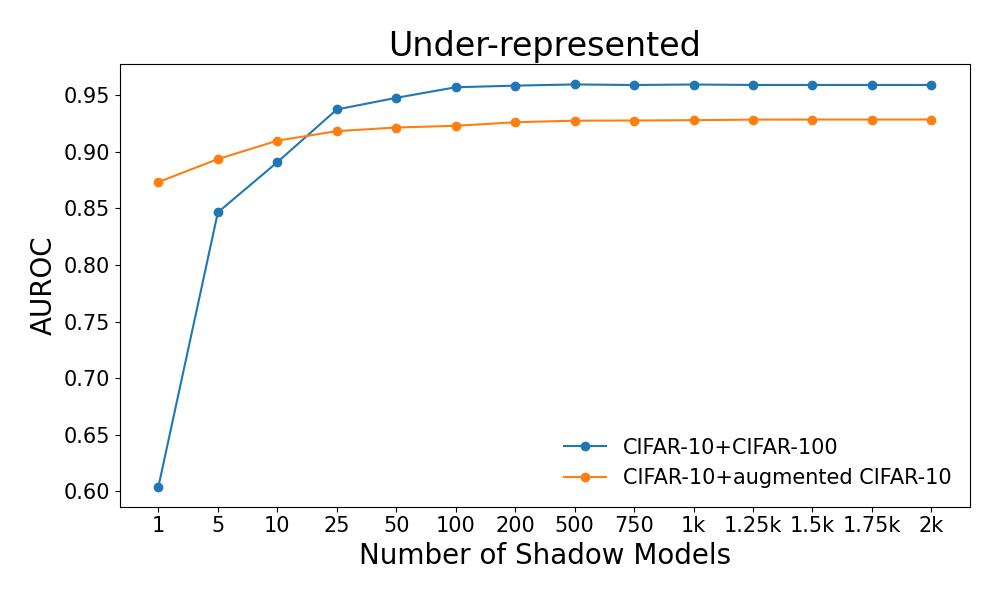}
    \caption{CIFAR-10 + OOD data}
    \label{fig:num_shadow_cifar10_ood_under}
  \end{subfigure}
  \begin{subfigure}{0.38\textwidth}
    \includegraphics[width=\linewidth]{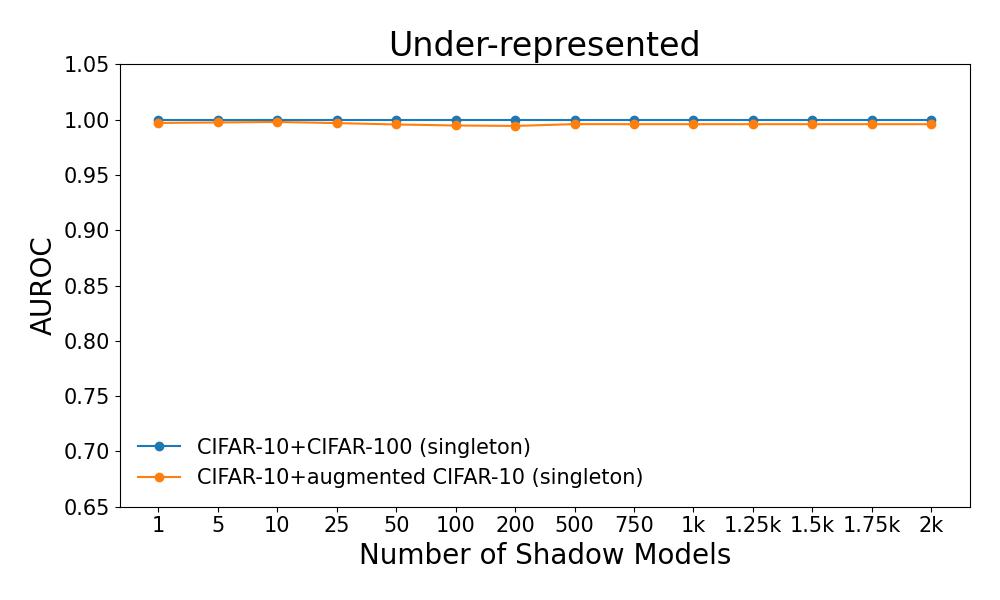}
    \caption{CIFAR-10 + HM OOD data}
    \label{fig:num_shadow_cifar10_mem_under}
  \end{subfigure}

  \caption{\textbf{AUROC  of MIA by Carlini \etal~\cite{carlini2022membership} varying the number of shadow models.} The adversary constructs dataset of different compositions, and picks challenge data from under-represented subpopulations. HM denotes highly memorized.
  }
  \label{fig:num_shadow_singleton}
\end{figure*}

%% file: sec_contents/8_future.tex
\section{Discussion}
\label{sec:discussion}

\noindent{\bf Connections between LiRA and Label Memorization:} The work of Carlini \etal~\cite{carlini2022membership} (LiRA) demonstrates state-of-the-art performance for MIAs, especially at low FPR regimes. To understand why, we urge the reader to analyze Algorithm 1 in their work. This approach is very similar to estimating memorization, as defined in Equation~\ref{eq:memorization}. The notable difference is line 4, where at each iteration, a dataset is sampled from the data distribution $\mathcal{D}$ (which does not happen for estimating memorization). However, upon analyzing their code~\footnote{\url{https://github.com/tensorflow/privacy/blob/master/research/mi_lira_2021/train.py\#L215:L228}}, we observed that the implementation is {\em not faithful} to the algorithm. In particular, the implementation draws random subsets of the {\em training data} (not the data distribution), and thus the LiRA attack {\em is equivalent to estimating} label memorization (with some additional hypothesis testing). The high success of their LiRA approach is yet another testament to our connection with memorization. 

\begin{figure}[t]
\begin{subfigure}{\linewidth}
\centering
\includegraphics[width=0.8\linewidth]{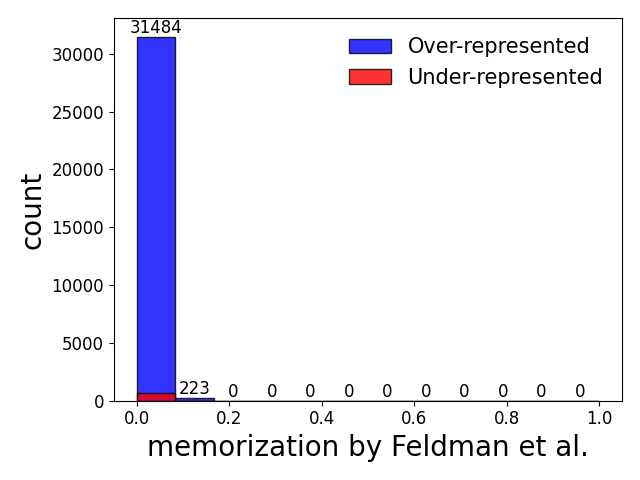}
\end{subfigure} 
\caption{Memorization scores obtained when training using DP-SGD on a dataset comprised of MNIST + SVHN.}
\label{fig:num_shadow_mnist_mem_under}
\end{figure}

\noindent{\bf Differential Privacy (DP) as a Defense:} DP is noted to be a promising defense against MIAs. We wish to study if this is the case for our "memorization-aware" attack as well. We use the exact same data (\ie same 1000 SVHN images out of the entire SVHN training data) and exact same subsampling (\ie 70\% out of the entire training set for every shadow model) as normal training cases in \S~\ref{sec:results}. We set target $\delta=10^{-5}$, train every shadow model for 100 epochs with noise multiplier $=1.3$, $\ell_2$ clipping norm $=1.0$, resulting $\varepsilon = 3.2$. Such a small privacy budget results in low utility models (which display poor generalization). From Figure~\ref{fig:num_shadow_mnist_mem_under}, we can see that when the model is trained with DP-SGD, samples that belong to the under-represented population also have low memorization scores (\ie samples are not memorized). Feldman~\cite{feldman2020longtail} notes that DP hampers memorization, and we observe a similar effect. 

\noindent{\bf Failed Inverse Generation Experiments:} Most of the MIAs attempt to ascertain distributional difference by training numerous shadow models, an expensive process. Both Brown \etal~\cite{brown2021memorization} and Feldman~\cite{feldman2020longtail} note that (a) models memorize samples that belong to low density subsets of the data manifold, and (b) memorization implies strong test performance on other samples from the same low density subset. To this end, we wished to design an MIA for highly memorized samples based on the aforementioned phenomenon. To this end, we trained a flow-based model~\cite{dinh2016realnvp} on the CIFAR-10 dataset. Inverse-flow models (denoted by functions $g^{-1}$ and $g$ s.t. $x = g \circ g^{-1}(x)$ for an input $x$) are able to find the corresponding latent vector (on the data manifold) for an input; we believed that such a model would be able to accurately identify the low density subset that a given sample belongs to. The procedure was as follows: for a sample $z=(x,y)$ that has a high memorization score, find the corresponding latent $l = g^{-1}(x)$. Then, we perturb the latent vector (\ie find other vectors within an $\delta$ threshold) to obtain a modified latent vector $l'$, which can converted to a modified input using the flow model \ie $x'=g(l')$. Ideally, evaluating the test accuracy of the model on $x'$ would help understand if $z$ is memorized \ie if accuracy is high, then w.h.p $z$ is memorized. Despite ensuring a performant inverse-flow model, this approach failed because we are unable to ascertain the ground truth label for the modified input \ie if label of $x$ is $y$, we are unable to prove that the label of $x'$ is also $y$.  While we believe this is a promising direction, and has been explored in the context of MIAs using generative adversarial networks~\cite{rezaei2022efficient}, more experimentation is needed before inverse-flow models can be used in conjunction with memorization for creating an MIA. 

\noindent{\bf Can OOD-ness be a Memorization Proxy?} Our proposal requires the existence of a memorization oracle $\mathcal{O}_{mem}$. Practically, this can be instantiated by running the algorithm specified by Feldman and Zhang~\cite{feldman2020neural}, which is a computationally expensive procedure. Recall that the results from \S~\ref{subsec:overheads} suggest that computational overheads can be reduced even for OOD samples. One naturally wonders if the degree of being an outlier (measured by the OOD score) can serve as a proxy for identifying samples with high memorization scores \ie are samples with high outlier scores those which are highly likely to be memorized? We plot the correlation between OOD-ness and memorization score in Figure~\ref{fig:mem-msp} and Figure~\ref{fig:mem-energy} in Appendix~\ref{app:additional-motivating-results}, and observe that no such correlation exists. This suggests that the characteristics captured by OOD detectors in the status quo are not the same as those captured by label memorization. Definitions that are tied to stability are more likely to capture this effect (refer Appendix~\ref{app:theory}). Thus designing inexpensive mechanisms to determine memorization scores remains an open problem.

\noindent{\bf The Privacy Onion Effect:} The definition of label memorization is implicitly dependent on the dataset (see Equation~\ref{eq:memorization}). 
Carlini \etal~\cite{carlini2022privacy} note that if a set of highly memorized points is removed from a dataset, points which previously had low memorization values now have high values (and vice versa). They term this the {\em privacy onion effect}. While defenders against memorization-guided MIAs may consider removing those samples that are likely to be memorized from the training dataset, two problems may emerge. The first is that the error of the model (on the low density subset associated with the deleted singleton) shoots up. Secondly, there will be a new set of points which are highly likely to be memorized. 



%% file: sec_contents/9_conclusions.tex
\section{Conclusion}
\label{sec:conclusion}

We attempt to explain the high efficacy of MIAs as a function of data's susceptibility to be memorized. 
We also experimentally verify our claims, and the results are in accordance with our expectations. More analysis is required to (a) convert the claims associated with mutual information from the work of Brown \etal~\cite{brown2021memorization} to probability bounds (needed to instantiate attack strategies), and (b) understand the efficacy of MI for samples that are unlikely to be memorized.

%% file: sec_contents/appendix/theory.tex
\section{Theoretical Extensions}
\label{app:theory}

We have the following lemma, which will come in handy in several scenarios.


\begin{lemma}
Let $g$ be any bounded function from $\Theta \times \mathcal{X}$ to $\mathbb{R}$. Then there exists an adversary whose advantage is equal to:
\[
\Pr_{\theta \in L(S \cup  \{ z \})} g(\theta,z) - \Pr_{\theta \in L(S) } g(\theta,z)
\]
Let us call the above expression $\delta (g,L,S,z)$. We immediately have.
\begin{eqnarray*}
    \mbox{Adv} (L,\mathcal{A}) & \geq & \delta(g,L,S,z)
\end{eqnarray*}
\end{lemma}

The proof structure is as follows: construct a probability distribution $p$ over $\Theta \times \mathcal{X}$ that is proportional to $f$ (\ie $p(\theta,z) = \alpha f(\theta,z)$.)). Consider an adversary that outputs $1$ with probability $p(\theta,z)$. Computing the advantage the proof is immediate. 

\noindent Similarly, the following items also hold
\begin{itemize}
\item[{\bf C1.}] Consider a bounded loss function $\ell$ with domain $\Theta \times \mathcal{X}$. Then we get the advantage:
\[
\Pr_{\theta \in L(S \cup  \{ z \})} \ell(\theta,z) - \Pr_{\theta \in L(S) } \ell(\theta,z)
\]
Note that the expression given corresponds to the stability of the loss function on a sample $z$ with respect to a dataset $S$.

\item[{\bf C2.}] Consider a bounded OOD detection function $ood$ with domain $\Theta \times \mathcal{X}$. Then we get the advantage:
\[
\Pr_{\theta \in L(S \cup  \{ z \})} ood(\theta,z) - \Pr_{\theta \in L(S) } ood(\theta,z)
\]
Note that the expression given corresponds to the stability of the OOD detection function on $z$ with respect to a dataset $S$.
\end{itemize}

%% file: sec_contents/appendix/implementation.tex
\section{Implementation Details}
\label{app:experimental}
We run all experiments with Tensorflow, Keras, Jax and NVDIA GeForce RTX 2080Ti GPUs. 
To accelerate the shadow model training when necessary, we use PyTorch with FFCV dataloader~\cite{leclerc2022ffcv}.

\section{Experimental Setting.}
\label{app:implementation-details}

\mypara{Datasets:} Here we describe the datasets used throughout our paper.
\begin{enumerate}
    \item {\bf MNIST}~\cite{lecun2010mnist}. The MNIST dataset consists of 60,000 handwritten digits (between 0 to 9) for train set, and 10,000 images for test set.
    \item {\bf SVHN}~\cite{svhn}. SVHN is the color images of real world house numbers. 
    We use the cropped version of the SVHN dataset, where images are tightly cropped around each digit of 0-9.
    From the original train/test set, we randomly select 1,000 images that are equally balanced across 10 labels. 
    \item {\bf CIFAR-10}~\cite{cifar10}. We use randomly sampled 6,000 images from the original 50,000 CIFAR-10 training images. We preserve their original labels so that images for class $i$ of CIFAR-10 where $i \in [0, 9]$ are assigned to the same class as the MNIST images with label $i$.
    \item {\bf CIFAR-100}~\cite{cifar10}. We use randomly sampled 1,000 images from the original train set of 50,000 CIFAR-100 images. We use the first 10 classes out of 100 classes in total including 100 images each.
    \item {\bf augMNIST and augCIFAR-10}. We generate the augmented variation of the MNIST dataset by using the technique in~\cite{hendrycks2022pixmix}.
    Original MNIST image is mixed with augmented versions of itself and the grayscale version of provided 14248 fractal images. 
    We use severity level of $2$ for mixing, and severity level $=4$ for base augmentation operations (such as normalization, random cropping, and rotation) for $4$ iterations for each image. 
    We use 1,000 images randomly chosen from the augmented train set, while ensuring the balance across 10 classes.
\end{enumerate}

\mypara{Models:} We adopt CNN architectures from Carlini \etal~\cite{carlini2022membership}: CNN models with 32 and 64 convolutional filters (referred to as CNN32 or CNN64, respectively).
We also include two ResNet architectures (ResNet-9 and Resnet-18)~\cite{resnet} into our consideration.
We use them when training shadow models or target models for MIAs.

\mypara{MIAs.} We briefly describle the details of different membership inference attack methods. 
\begin{enumerate}
    \item Yeom \etal~\cite{yeom2018privacy}. The attack is based on the observation that ML models are trained to minimize the loss of the training samples, and the loss values on them are more likely to be lower than the samples outside of the training set.
    They suggest applying threshold on the loss values from the ML model to infer the membership of an input.
    \item Shokri \etal~\cite{shokri2017membership}. The attack uses a trained ML model to ascertain membership/non-membership. In our experiments, a fully-connected neural network with one hidden layer of size 64 with ReLU (rectifier linear units) activation functions and a SoftMax layer is used to distinguish feature vectors obtained from shadow models trained with and without a data-point.
    \item Song \etal~\cite{song2021systematic}. The attack uses shadow models to approximate the distributions of entropy values, instead of cross-entropy loss. Given a target model and sample, they conduct hypothesis test between the member and non-member distributions for each class.
    \item Sablayrolles \etal~\cite{sablayrolles2019white}. Their attack also utilizes the loss value. The loss is scaled for better attack accuracy, using a per-sample hardness threshold which is identified in a non-parametric way from shadow model training.  
    \item Carlini \etal~\cite{carlini2022membership}. Refer Algorithm~\ref{alg:mem-privacy} in Appendix~\ref{app:algorithm}. We used the implementation provided by the authors for the attack.
\end{enumerate}

\input{sec_contents/appendix/algorithm}

%% file: sec_contents/appendix/algorithm.tex
\subsection{Algorithm for MIA and Memorization}
\label{app:algorithm}

We describe the algorithm proposed by Carlini \etal~\cite{carlini2022membership} for MI (reproduced from their work), and Feldman and Zhang~\cite{feldman2020neural} for label memorization in Algorithm~\ref{alg:mem-privacy}. Observe that both the MI success and label memorization can be obtained from the same algorithm by saving some state. 

\noindent{\bf Common Subroutines:} Notice that the privacy score estimation relies on training numerous models, some with the point under consideration, and some without (refer the grey box in Algorithm~\ref{alg:mem-privacy}). We refer to this as the leave-one-out (LOO) subroutine. Observe that the exact same LOO subroutine can be used to empirically estimate memorization (which is also used to measure algorithmic stability and influence). As noted in \S~\ref{sec:discussion}, the attack by Carlini \etal~\cite{carlini2022membership} couples this with hypothesis testing, leading to high success rates for all samples. 



\begin{algorithm}[h]
 \begin{algorithmic}[1]
  \REQUIRE \text{model} $\theta$, \text{example} $z=(x, y)$, \text{data distribution} $\mathcal{D}$
  \STATE $\text{confs}_{\text{in}} = \{\}$
  \STATE $\text{confs}_{\text{out}} = \{\}$
  \STATE $\mathcal{F}_{\text{in}} = \{\}$
  \STATE $\mathcal{F}_{\text{out}} = \{\}$
\begin{tcolorbox}
  \FOR{$m$ times}
    \STATE $\hat{S} \gets^\$ \mathcal{D}$ \algcomment{Sample a shadow dataset}
    \STATE $\theta_{\text{in}} \gets L(\hat{S} \cup \{(x,y)\})$ \algcomment{train IN model}
    \STATE $\mathcal{F}_{\text{in}} \gets \mathcal{F}_{\text{in}} \cup \{\theta_{\text{in}}\}$
    \STATE $\theta_{\text{out}} \gets L(\hat{S} {\setminus \{(x,y)\}})$ \algcomment{train OUT model}
    \STATE $\mathcal{F}_{\text{out}} \gets \mathcal{F}_{\text{out}} \cup \{\theta_{\text{out}}\}$
    \STATE $\text{confs}_{\text{in}} \gets \text{confs}_{\text{in}} \cup \{\phi(\theta_{\text{in}}(x)_y)\}$
    \STATE $\text{confs}_{\text{out}} \gets \text{confs}_{\text{out}} \cup \{\phi(\theta_{\text{out}}(x)_y)\}$
  \ENDFOR
\end{tcolorbox}  
  \STATE $\mu_{\text{in}} \gets \texttt{mean}(\text{confs}_{\text{in}})$
  \STATE $\mu_{\text{out}} \gets \texttt{mean}(\text{confs}_{\text{out}})$
  \STATE $\sigma_{\text{in}}^2 \gets \texttt{var}(\text{confs}_{\text{in}})$
  \STATE $\sigma_{\text{out}}^2 \gets \texttt{var}(\text{confs}_{\text{out}})$
   \STATE $\text{confs}_{\text{obv}} \gets \theta(x)_y$
  \vspace{0.5em}
  \STATE $\Lambda = \frac{p(\text{confs}_{\text{obv}}|\mathcal{N}(\mu_{\text{in}}, \sigma_{\text{in}}))}{p(\text{confs}_{\text{obv}})|\mathcal{N}(\mu_{\text{out}}, \sigma_{\text{out}}))}$
  \STATE $\widetilde{\mem}(\mathcal{A},S,z) := \Pr_{\theta_{\text{in}} \in \mathcal{F}_{\text{in}}}[\theta_{\text{in}}(x) = y] - \Pr_{\theta_{\text{out}} \in \mathcal{F}_{\text{out}}}[\theta_{\text{out}}(x) = y]$ 
  \vspace{0.5em}
  \RETURN $\frac{|\mu_{\text{in}} - \mu_{\text{out}}|}{\sigma_{\text{in}} + \sigma_{\text{out}}}$~ and ~$\widetilde{\mem}(\mathcal{A},S,z)$~ and ~$\Lambda$
  \end{algorithmic}
 \caption{\textbf{Algorithm for estimating memorization, privacy score, and the outcome of Carlini \etal~\cite{carlini2022privacy} attack.} 
 }
 \label{alg:mem-privacy}
\end{algorithm}

%% file: sec_contents/appendix/additional_results.tex
\section{Additional Results}
\label{app:additiona-results}

\subsection{Motivating results}
\label{app:additional-motivating-results}
\begin{figure}[bht]
\centering
\begin{subfigure}{0.32\columnwidth}
\centering
\includegraphics[width=\textwidth]{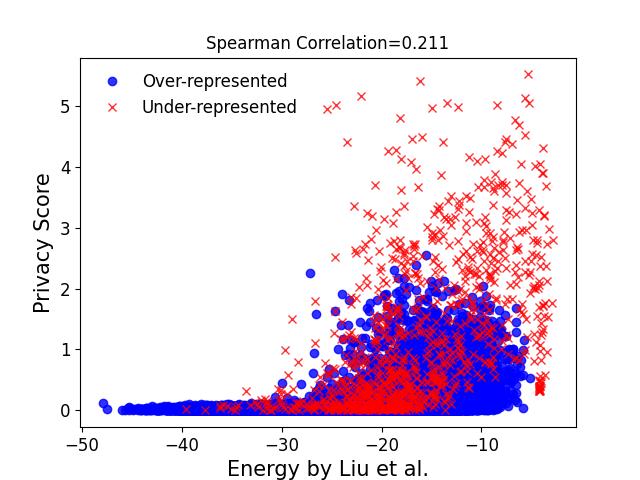}
\caption{augMNIST ({\bf V1})}
\end{subfigure}
\begin{subfigure}{0.32\columnwidth}
\centering
\includegraphics[width=\textwidth]{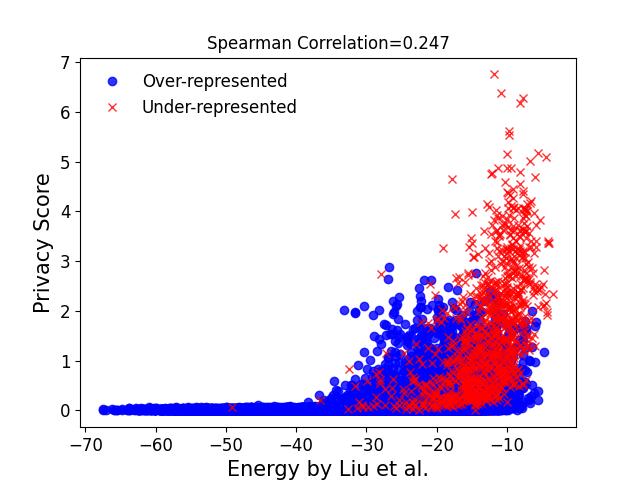}
\caption{SVHN ({\bf V2})}
\end{subfigure}
\begin{subfigure}{0.32\columnwidth}
\centering
\includegraphics[width=\textwidth]{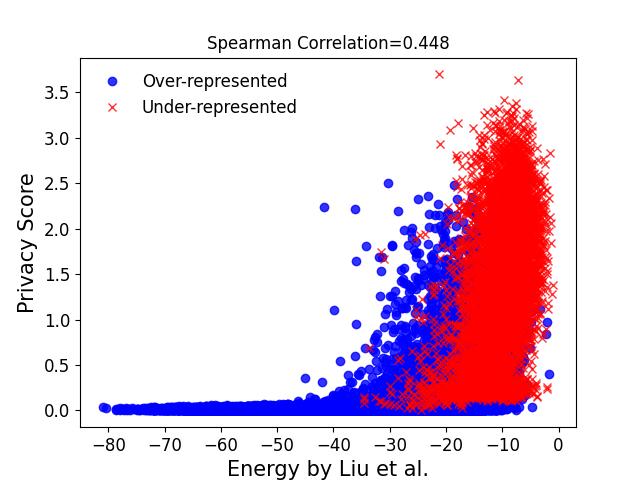}
\caption{CIFAR-10 ({\bf V3})}
\end{subfigure}
\caption{\textbf{OOD data is not always more susceptible to MI attack than ID data.} We plot the privacy score distribution (a) \textbf{L}: CNN32 trained on MNIST+augMNIST (b) \textbf{M}: CNN32 trained on MNIST+SVHN (c) \textbf{R}: CNN64 trained on MNIST+CIFAR-10.
}
\label{fig:energy-privacy-mnist}
\end{figure}

Refer to Figure~\ref{fig:energy-privacy-mnist} along with Figure~\ref{fig:privacy-ood-mnist}.

\begin{table*}[th]
\resizebox{\textwidth}{!}{
\centering
\footnotesize
\begin{tabular}{ll|cc|lcccc}
\toprule
\multirow{2}{*}{\bf{Over-represented}} & \multirow{2}{*}{\bf{Under-represented}} & \multicolumn{2}{c|}{\bf \# Samples} & \multirow{2}{*}{\bf Model} & \multicolumn{2}{c}{\bf Train Acc. (\%)} & \multicolumn{2}{c}{\bf Test Acc. (\%)} \\ \cline{6-9}
  & & $|\Dotr|$ & $|\Dutr|$ & & $\Dotr$ & $\Dutr$ & $\Dote$ & $\Dute$ \\ 
\hline\hline
\multirow{3}{*}{MNIST}&augmented MNIST & 60,000 & 219 & CNN32 & 99.27 $\pm$ 0.14 & 72.08 $\pm$ 5.93 & 98.44 $\pm$ 0.20 & 49.38 $\pm$ 3.04\\
&cropped SVHN & 60,000 & 209 & CNN32 & 99.32 $\pm$ 0.10 & 71.40 $\pm$ 6.22 & 98.48 $\pm$ 0.20 & 19.67 $\pm$ 2.82 \\
&CIFAR-10 & 60,000 & 2153 & CNN64 & 99.43 $\pm$ 0.35 & 72.66 $\pm$ 1.86 & 98.28 $\pm$ 0.31 & 20.78 $\pm$ 1.215\\ \cdashline{1-9}
\multirow{1}{*}{CIFAR-10} & augmented CIFAR-10 & 50,000 & 274 & ResNet-18 & 96.35 $\pm$ 0.84 & 71.02 $\pm$ 6.21 & 87.83 $\pm$ 1.57 & 50.13 $\pm$ 2.02 \\
& CIFAR-100 & 50,000 & 254 & ResNet-9 & 96.61 $\pm$ 0.2 & 69.69 $\pm$ 6.1 & 87.35 $\pm$ 0.56 & 11.21 $\pm$ 1.6 \\ \cline{1-9}
\end{tabular}}
\caption{{\bf Salient features of the datasets used in our experiments.} Accuracy reported in $95\%$ confidence interval over 2000 iterations as in Table~\ref{tab:salient}.}
\label{tab:salient-singleton}
\end{table*}

\subsection{Attacks}
\label{app:additional-results-mia}

\begin{table*}[h]
\begin{adjustbox}{width=\columnwidth,center}
\begin{tabular}{l|l|cc|cc}
\toprule
\multirow{2}{*}{\bf Method} & \multirow{2}{*}{{\bf Dataset} ($\Dtr$)} & \multicolumn{2}{c|}{{\bf AUROC} $\uparrow$} & \multicolumn{2}{c}{{\bf TPR @ $0.1 \%$ FPR} $\uparrow$} \\ \cline{3-6}
                                & \multicolumn{1}{c|}{} & \multicolumn{1}{c}{All} & Under-represented    & \multicolumn{1}{c}{All} & Under-represented \\ 
\hline \hline
\multirow{3}{*}{Yeom \etal~\cite{yeom2018privacy}} & MNIST+CIFAR-10 (random)& 0.51 & 0.73 & 0.0 \% & 0.0 \% \\
                                & MNIST+CIFAR-10 (subgroup)& 0.50 & 0.80 & 0.0 \% & 0.0 \%\\
                                & MNIST+CIFAR-10 (singletons) & 0.51 & \textbf{0.96} & 0.0 \% & 0.0 \% \\
                                \cline{1-6}
                                
\multirow{3}{*}{Shokri \etal~\cite{shokri2017membership}} & MNIST+CIFAR-10 (random)& 0.56 & 0.85 & 0.66 \% & 1.09 \% \\
                                & MNIST+CIFAR-10 (subgroup) & 0.52 & 0.86 & 0.16 \% & 0.98 \% \\
                                & MNIST+CIFAR-10 (singletons) & 0.55  & \textbf{0.99} & 0.95 \% & \textbf{16.52 \%} \\
                                \cline{1-6}
\multirow{3}{*}{Sablayarolles \etal~\cite{sablayrolles2019white}} & MNIST+CIFAR-10 (random)& 0.54 & 0.87 & 0.20 \% & 1.98 \% \\
                                 & MNIST+CIFAR-10 (subgroup)& 0.53 & 0.92 & 1.22 \% & 14.95 \% \\
                                & MNIST+CIFAR-10 (singletons) & 0.53 & \textbf{1.0} & 1.98 \% & \textbf{21.50 \%} \\
                                \cline{1-6}

\multirow{3}{*}{Song \etal~\cite{song2021systematic}} & MNIST+CIFAR-10 (random)& 0.54 & 0.87 & 0.20 \% & 2.06 \% \\
                                & MNIST+CIFAR-10 (subgroup)& 0.53 & 0.95 & 0.20 \% & 6.09 \% \\
                                & MNIST+CIFAR-10 (singletons) & 0.53 & \textbf{0.99} & 1.98 \% & \textbf{15.53 \%} \\
                                \cline{1-6}                                
                                
\multirow{3}{*}{Carlini \etal~\cite{carlini2022membership}} & MNIST+CIFAR-10 (random)& 0.59 & 0.96 & 5.7 \% & 44.15 \% \\
                                & MNIST+CIFAR-10 (subgroup)& 0.52 & 0.97 & 0.27 \% & 32.44 \% \\
                                & MNIST+CIFAR-10 (singletons) & 0.56 & \textbf{1.0} & 4.58 \% & \textbf{96.51 \%} \\ \cline{1-6}
\end{tabular}%
\end{adjustbox}
\caption{\textbf{Additional comparison of representative MIAs.} 
For each choice of MIA and mixture dataset, we direct readers to compare AUROC and TPR between (i) (All) vs. (Under-represented), and (ii) (random) vs. (subgroup) vs. (singletons) to see the effect of utilizing singletons in MIAs. Best results are \textbf{boldfaced}.}
\label{tab:mia-results-mnist-additional}
\end{table*}

\begin{figure}[thb]
\centering
\begin{subfigure}{0.42\columnwidth}
\centering
\includegraphics[width=\linewidth]{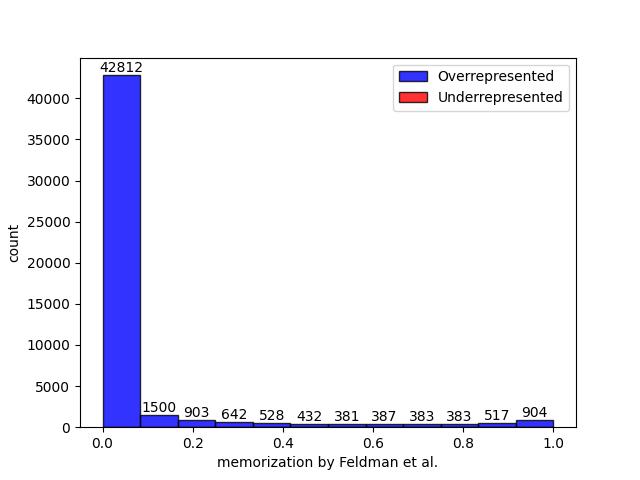}
\caption{CIFAR-10}
\label{fig:mem-distr-cifar10}
\end{subfigure}
\begin{subfigure}{0.42\columnwidth}
\centering
\includegraphics[width=\linewidth]{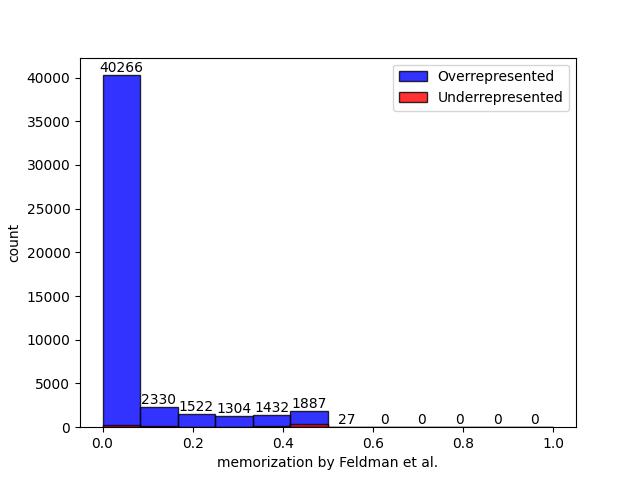}
\caption{CIFAR-10+CIFAR-100}
\label{fig:mem-distr-cifar10+cifar100}
\end{subfigure}
\caption{\textbf{Distribution of memorization scores.}}
\label{fig:varun}
\end{figure}

\begin{figure}[htp]
  \centering
  

  \begin{subfigure}{0.47\columnwidth}
    \includegraphics[width=\linewidth]{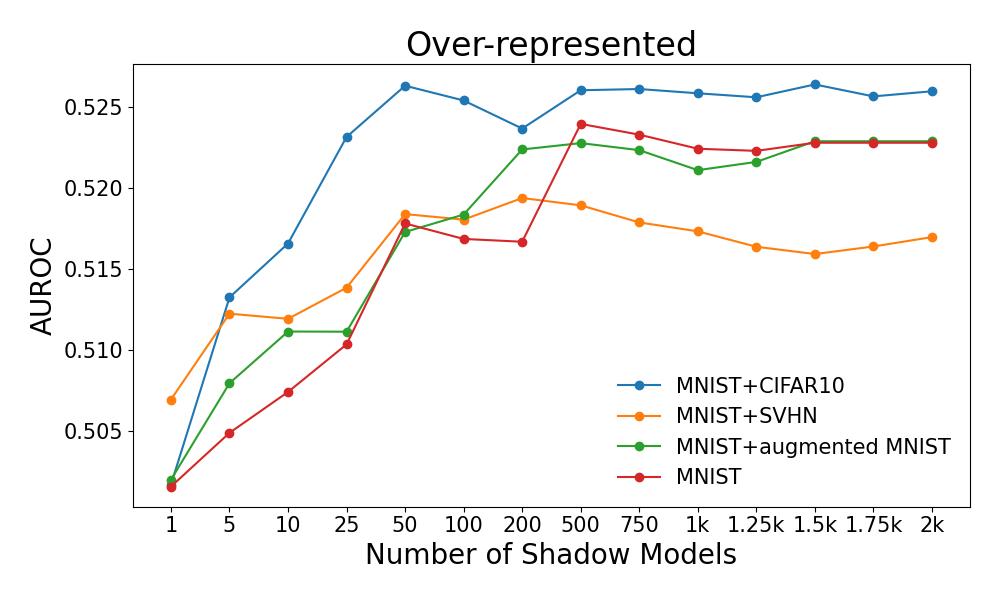}
    \caption{MNIST+random OOD}
    \label{fig:num_shadow_mnist_ood_over}
  \end{subfigure}
  \begin{subfigure}{0.47\columnwidth}
    \includegraphics[width=\linewidth]{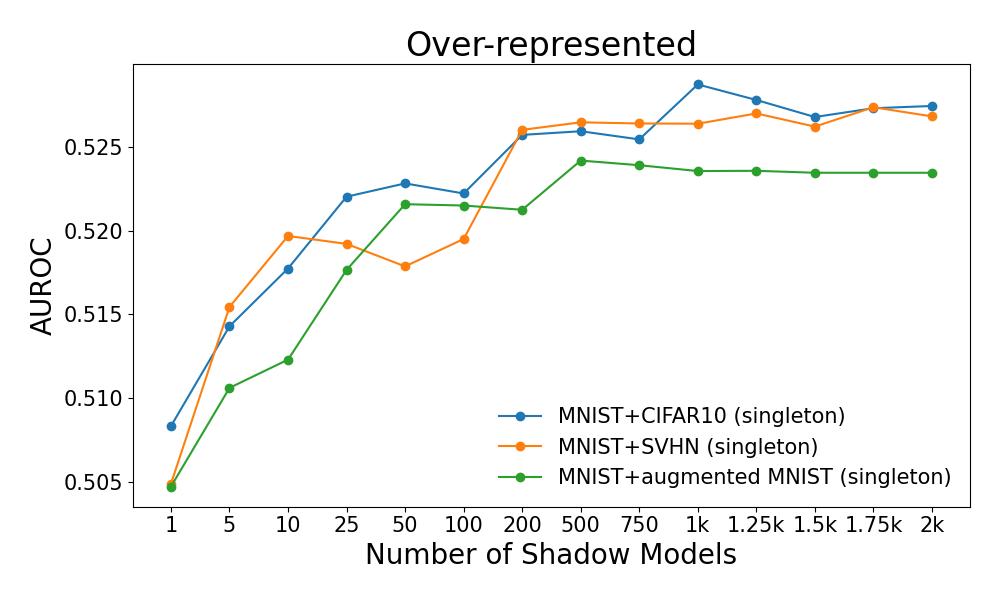}
    \caption{MNIST+HM OOD}
    \label{fig:num_shadow_mnist_mem_over}
  \end{subfigure} 
  \\
  \begin{subfigure}{0.47\columnwidth}
    \includegraphics[width=\linewidth]{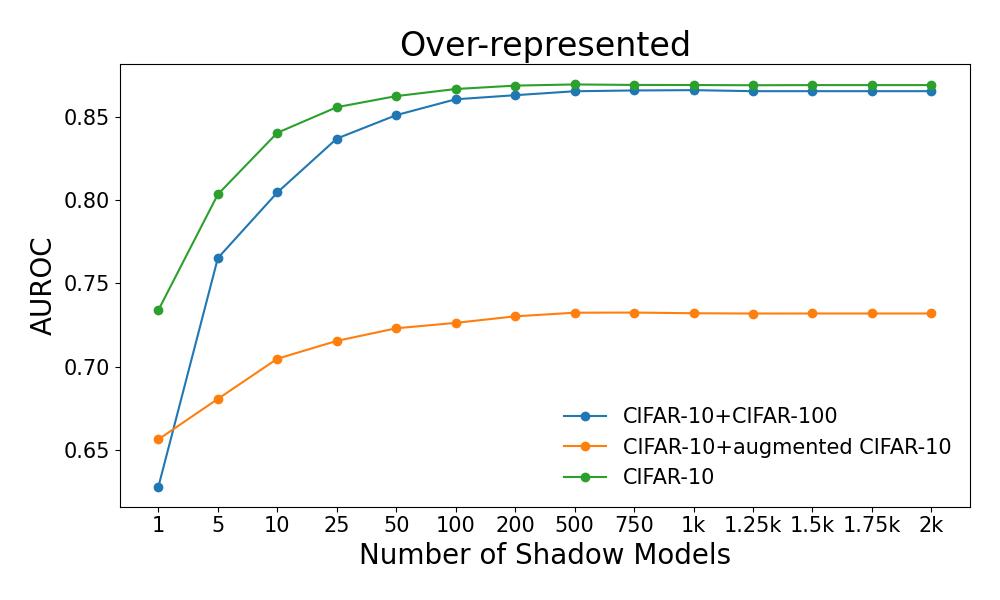}
    \caption{CIFAR-10+random OOD}
    \label{fig:num_shadow_cifar10_ood_over}
  \end{subfigure}
  \begin{subfigure}{0.47\columnwidth}
    \includegraphics[width=\linewidth]{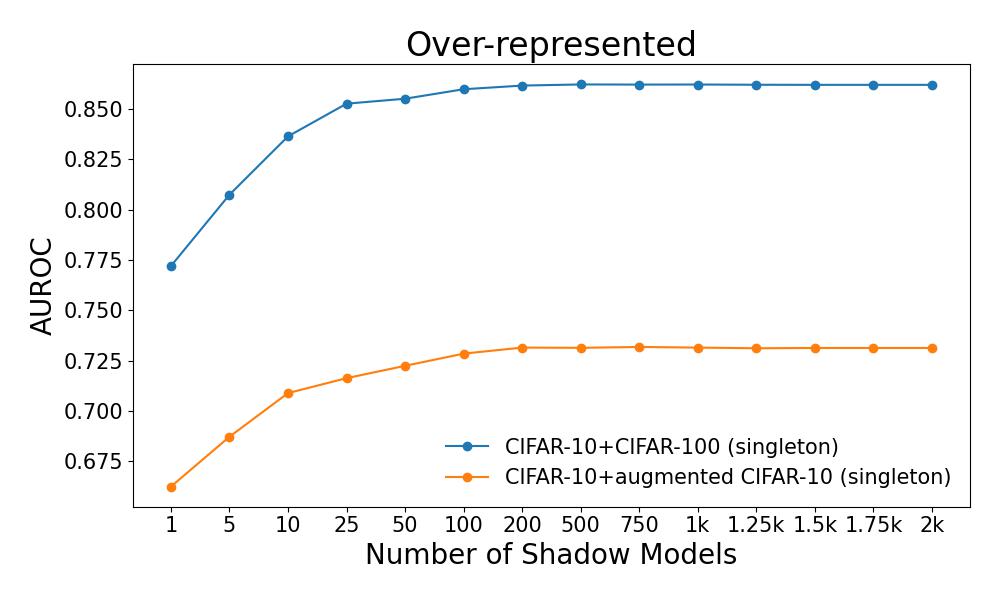}
    \caption{CIFAR-10+HM OOD}
    \label{fig:num_shadow_cifar10_mem_over}
  \end{subfigure}

  \caption{\textbf{AUROC of MIA by Carlini \etal~\cite{carlini2022membership} varying the number of shadow models.} Adversary picks challenge data from over-represented subpopulation. HM stands for ``Highly Memorized.''}
  \label{fig:num_shadow_overrepresented}
\end{figure}

When using random MNIST data, over 1000 shadow models are needed, and more than 100 shadow models are required for random CIFAR-10 data (see Figure~\ref{fig:num_shadow_mnist_ood_over} and Figure~\ref{fig:num_shadow_cifar10_ood_over})
Compare it with reduction in the number of required shadow models observed in Figure~\ref{fig:num_shadow_singleton}.

\begin{figure}[h]
  \centering
  

  \begin{subfigure}{0.47\columnwidth}
    \includegraphics[width=\linewidth]{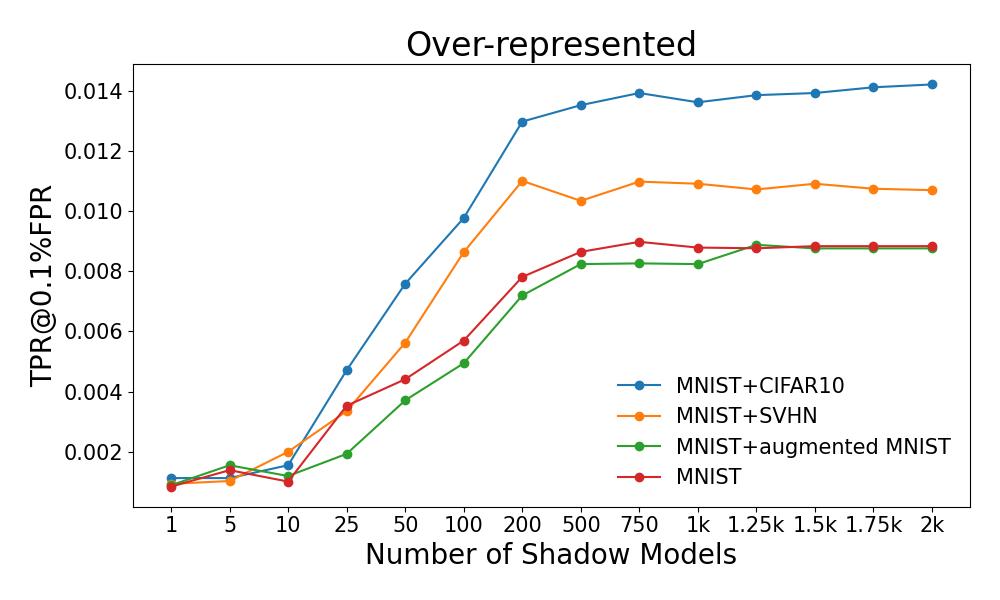}
    \caption{MNIST+ramdom OOD, attack on random data}
  \end{subfigure}
  \begin{subfigure}{0.47\columnwidth}
    \includegraphics[width=\linewidth]{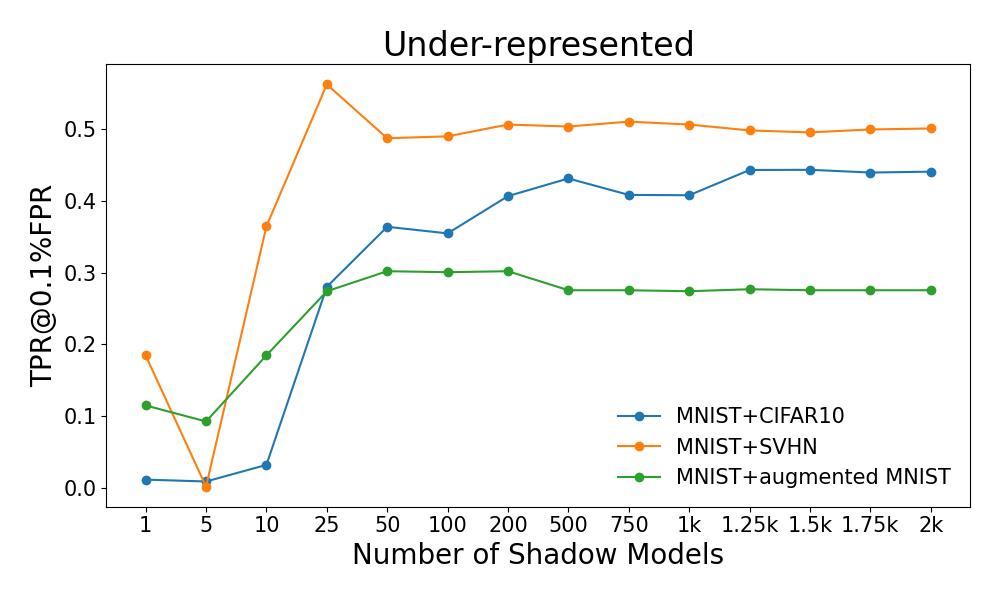}
    \caption{MNIST+random OOD, attack on HM OOD data}
  \end{subfigure} 
  \\
  \begin{subfigure}{0.47\columnwidth}
    \includegraphics[width=\linewidth]{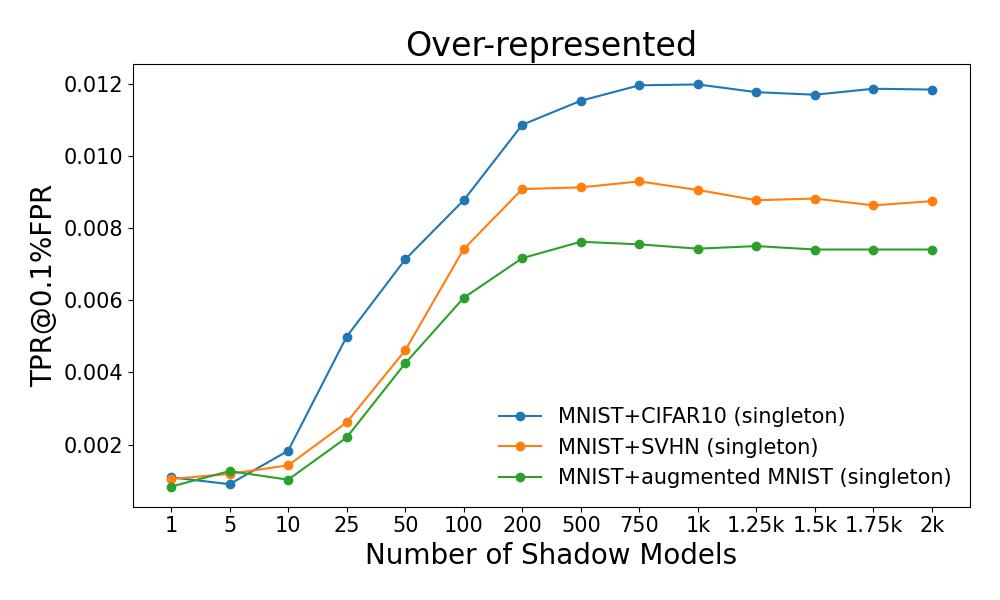}
    \caption{MNIST+HM OOD, attack on random data}
  \end{subfigure}
  \begin{subfigure}{0.47\columnwidth}
    \includegraphics[width=\linewidth]{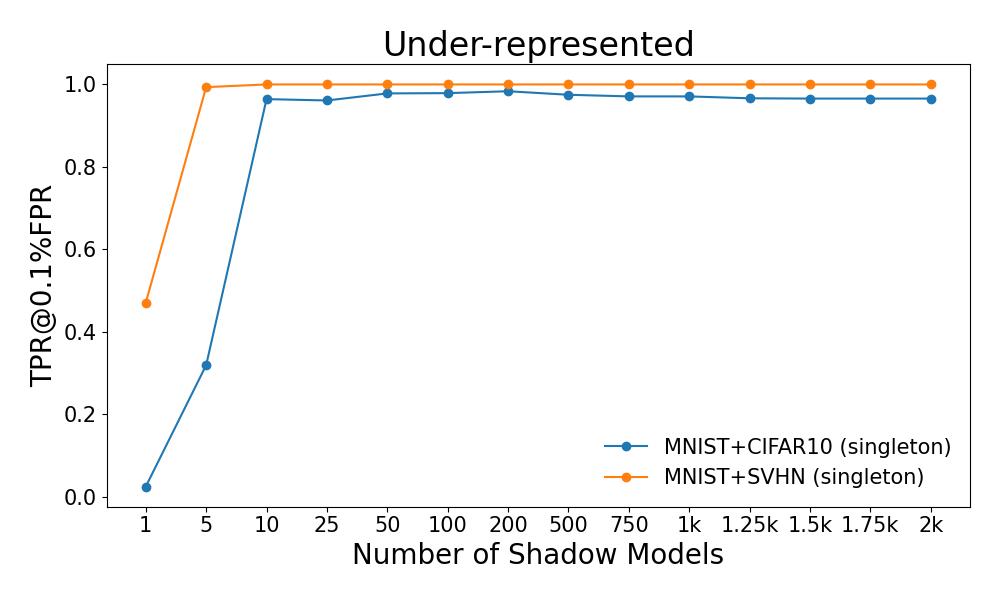}
    \caption{MNIST+HM OOD, attack on HM OOD data}
  \end{subfigure}

  \caption{\textbf{TPR at FPR$=0.1 \%$ of MIA by Carlini \etal~\cite{carlini2022membership} varying the number of shadow models.} 
  Challenge data is either from over-represented subpopulations, or HM samples. HM stands for ``Highly Memorized.''}
  \label{fig:num_shadow_mnist_tpr}
\end{figure}

\input{sec_contents/new_correlation_figs}

%% file: sec_contents/new_correlation_figs.tex
\begin{figure*}[bht]
\centering
\begin{subfigure}{0.32\textwidth}
\centering
\includegraphics[width=\textwidth]{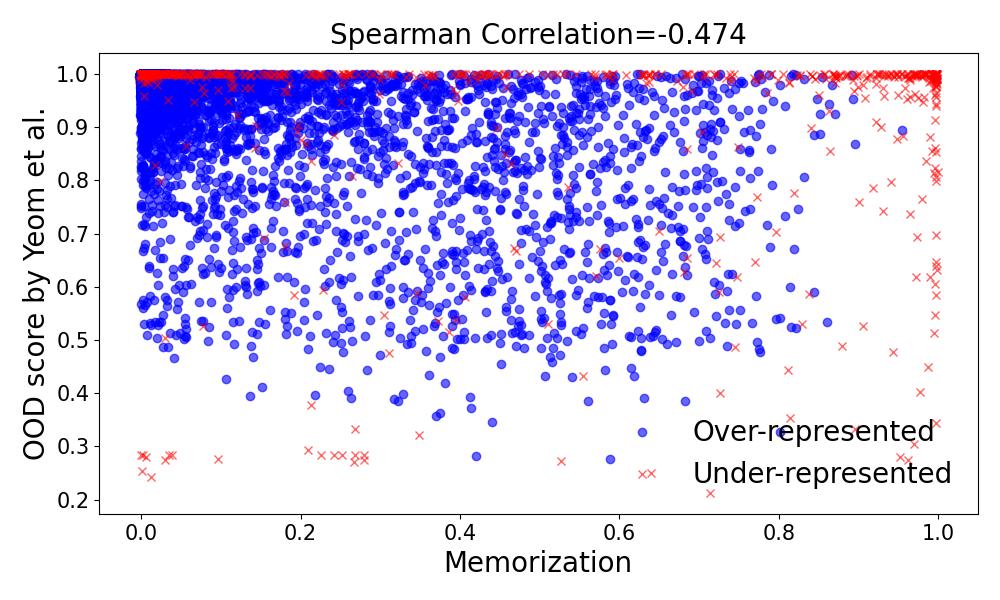}
\caption{MNIST + augMNIST}
\includegraphics[width=\textwidth]{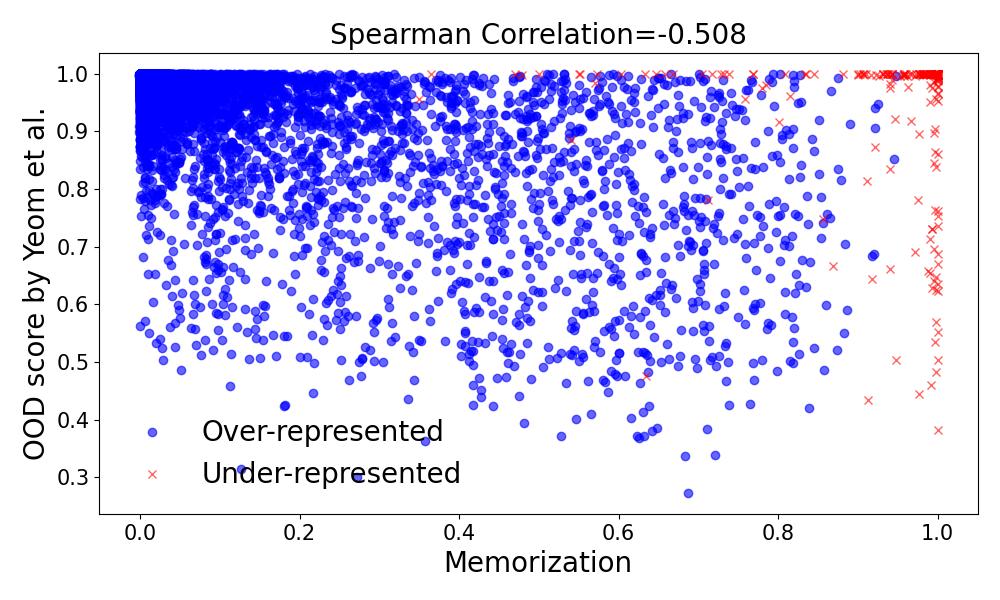}
\caption{MNIST + HM augMNIST}
\end{subfigure}
\begin{subfigure}{0.32\textwidth}
\centering
\includegraphics[width=\textwidth]{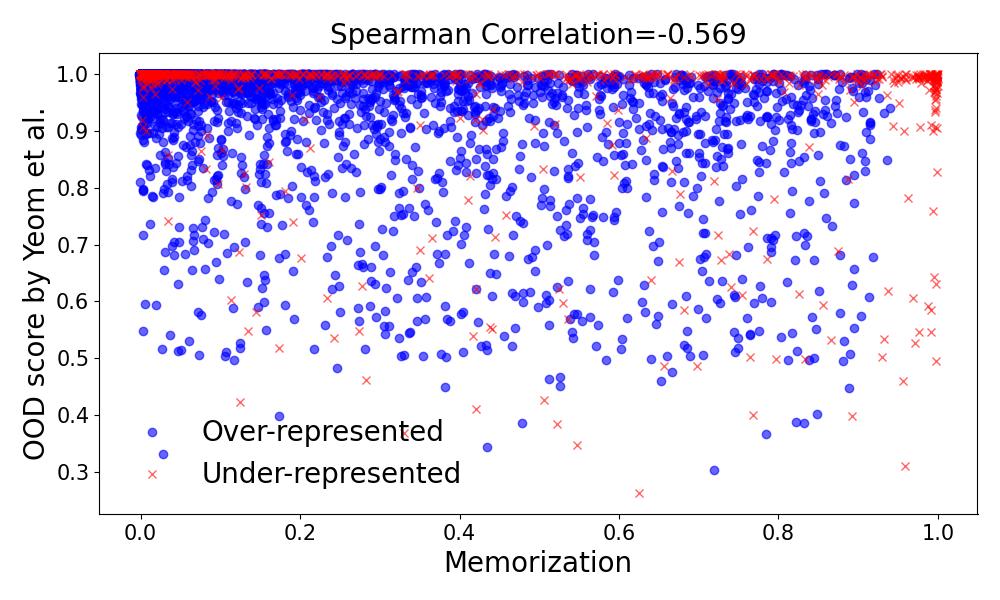}
\caption{MNIST + SVHN}
\includegraphics[width=\textwidth]{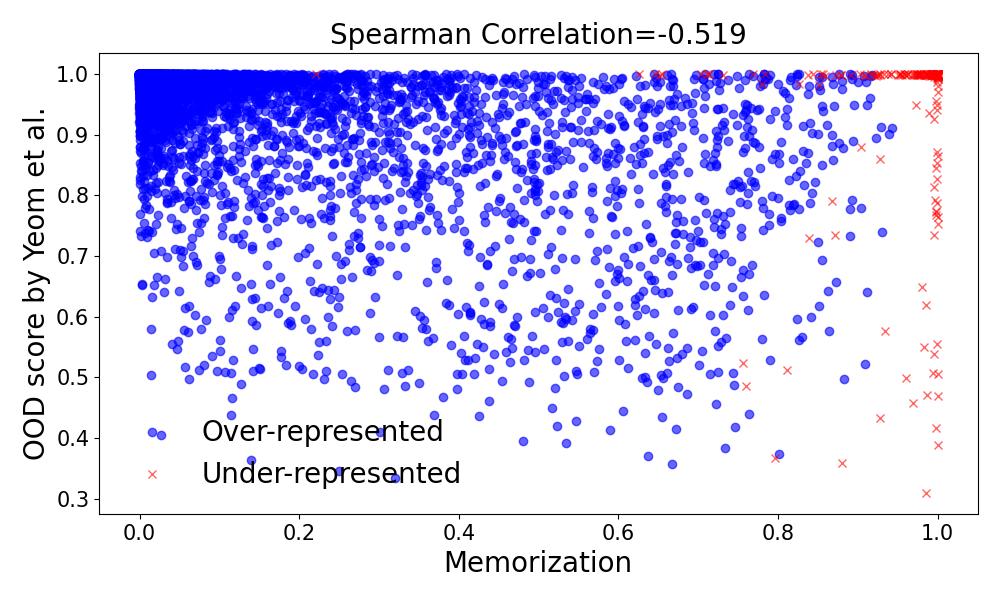}
\caption{MNIST + HM SVHN}
\end{subfigure}
\begin{subfigure}{0.32\textwidth}
\centering
\includegraphics[width=\textwidth]{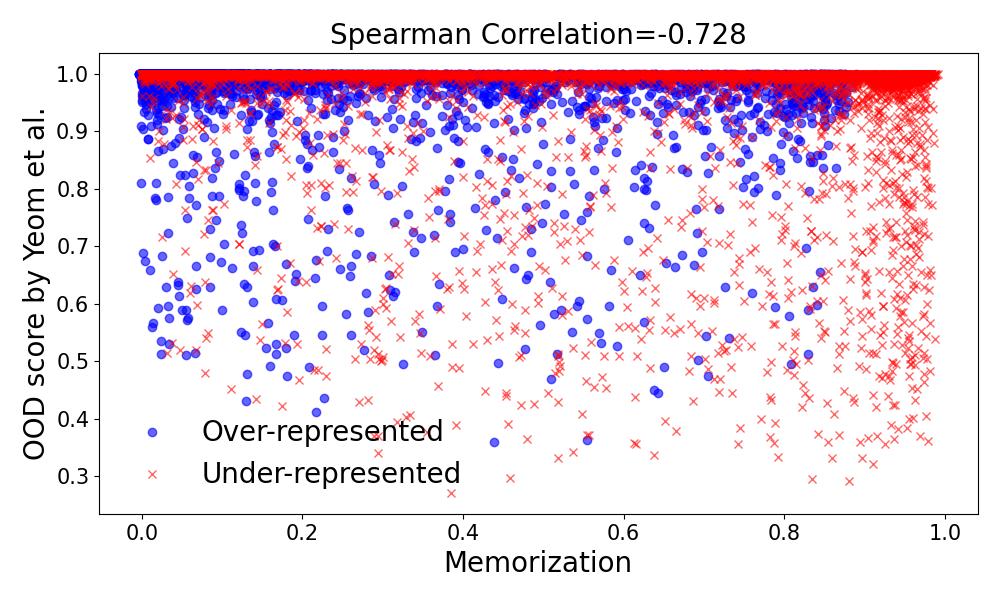}
\caption{MNIST + CIFAR-10}
\includegraphics[width=\textwidth]{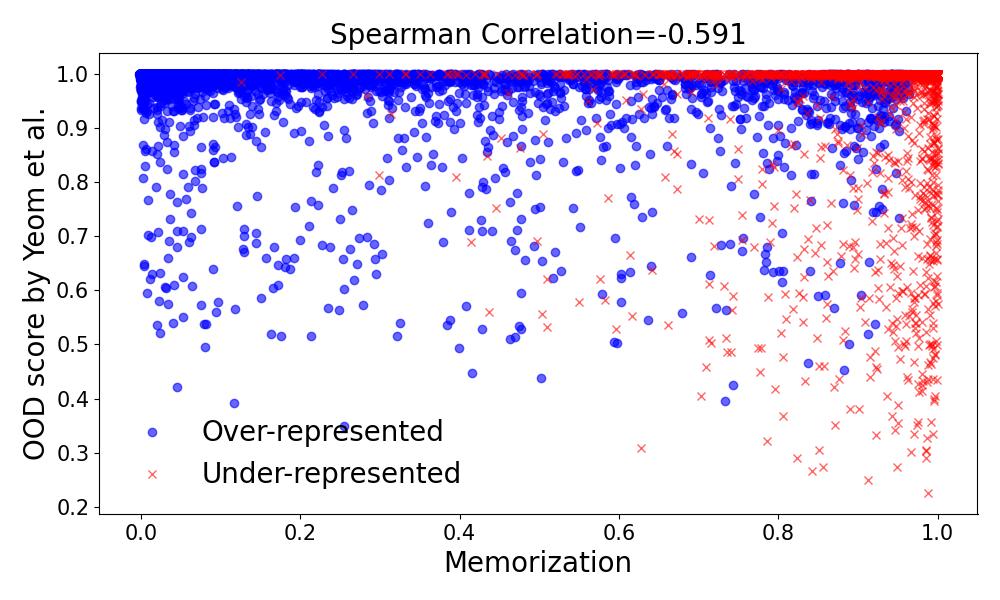}
\caption{MNIST + HM CIFAR-10}
\end{subfigure}
\caption{\textbf{Memorization vs. OOD score by Hendrycks \etal~\cite{hendrycks2016msp}.} Adversary constructs the dataset varying the choice of samples for the under-represented population. ``HM'' refers to Highly Memorized. We observe clearer separation between the populations (blue vs. red) in the second row.}
\label{fig:mem-msp}
\end{figure*}

\begin{figure*}[bht]
\centering
\begin{subfigure}{0.32\textwidth}
\centering
\includegraphics[width=\textwidth]{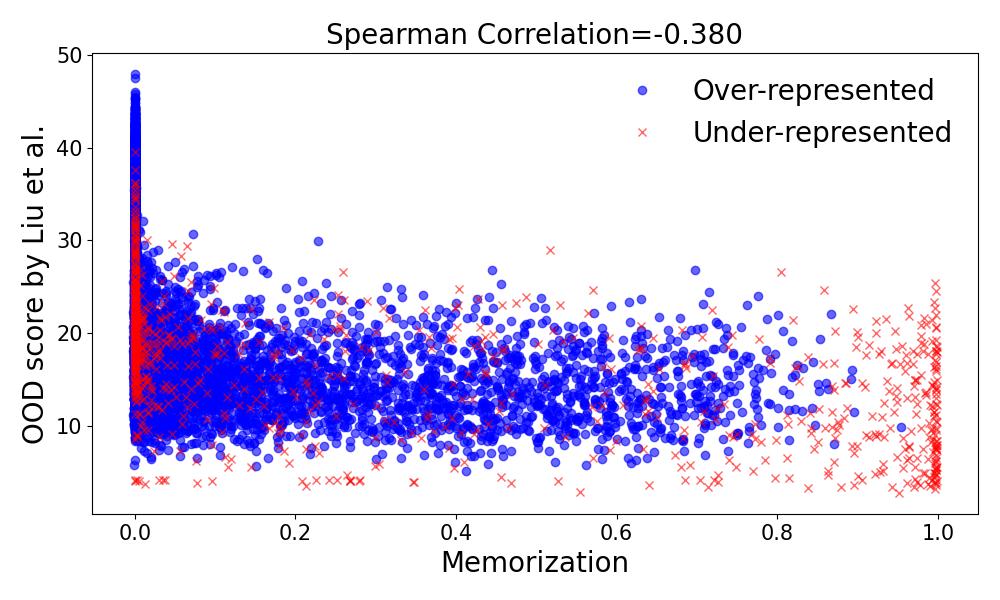}
\caption{MNIST+augmented MNIST}
\includegraphics[width=\textwidth]{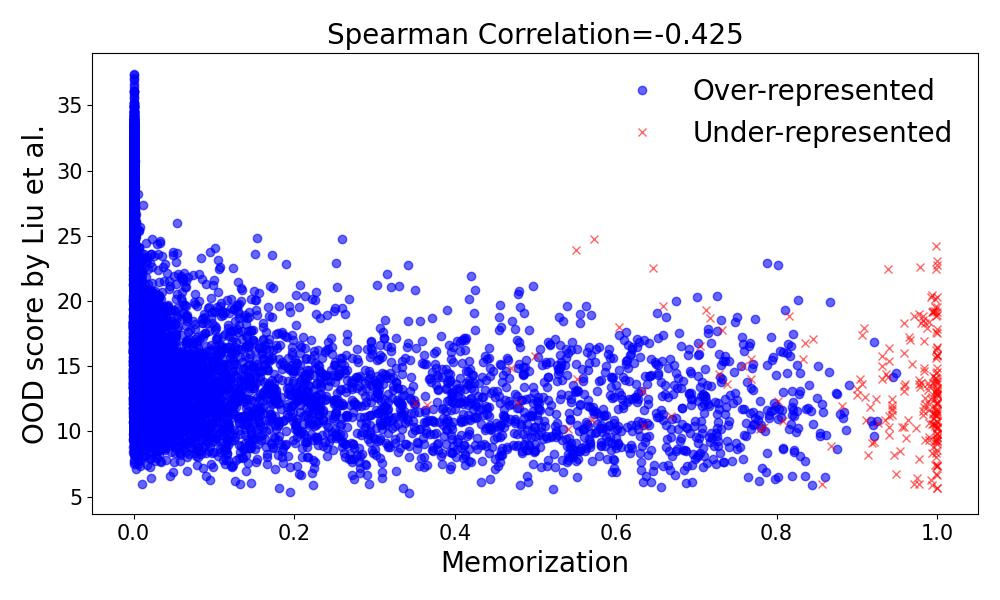}
\caption{MNIST+HM augmented MNIST}
\end{subfigure}
\begin{subfigure}{0.32\textwidth}
\centering
\includegraphics[width=\textwidth]{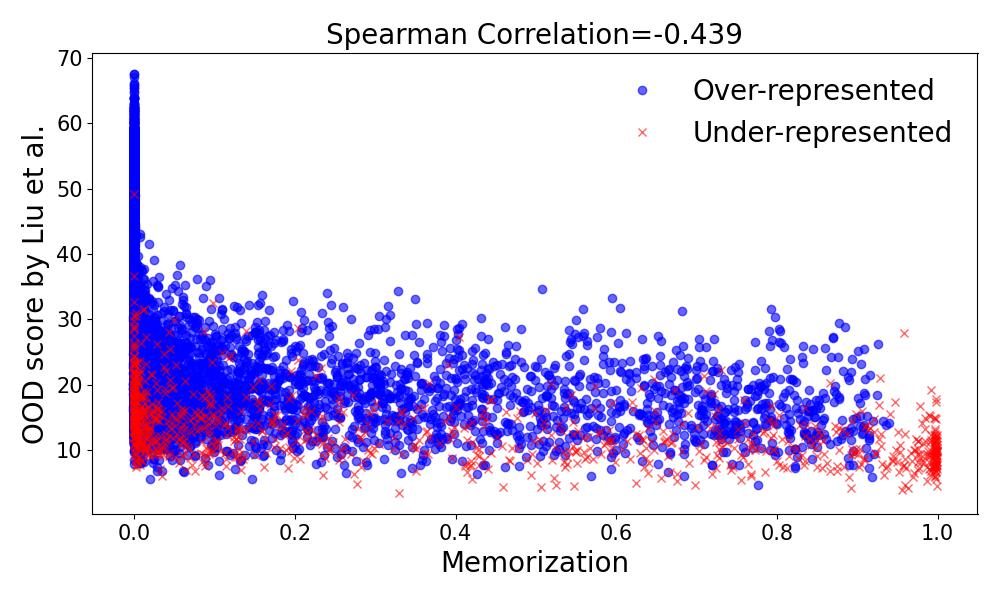}
\caption{MNIST+SVHN}
\includegraphics[width=\textwidth]{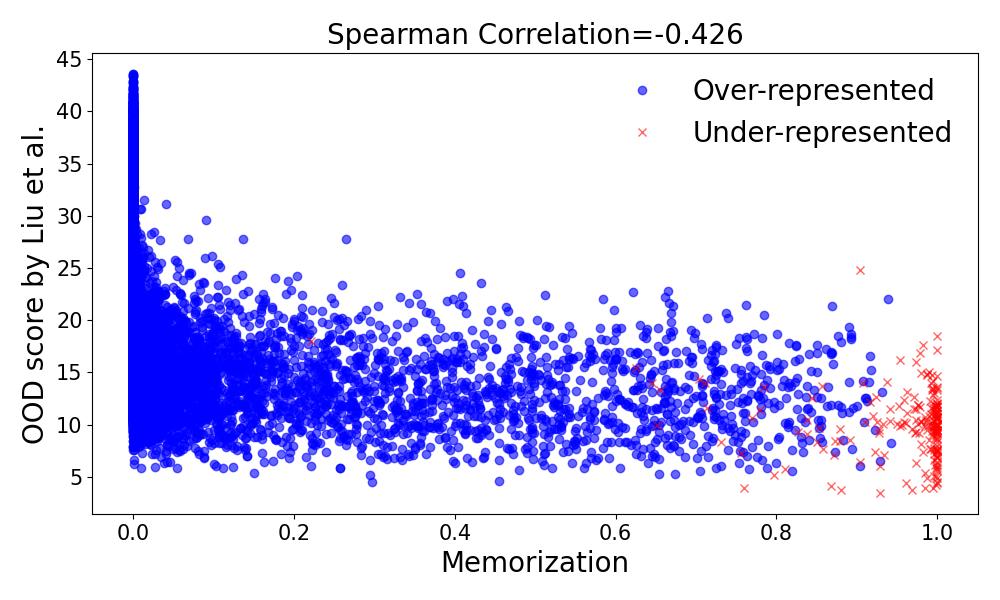}
\caption{MNIST+HM SVHN}
\end{subfigure}
\begin{subfigure}{0.32\textwidth}
\centering
\includegraphics[width=\textwidth]{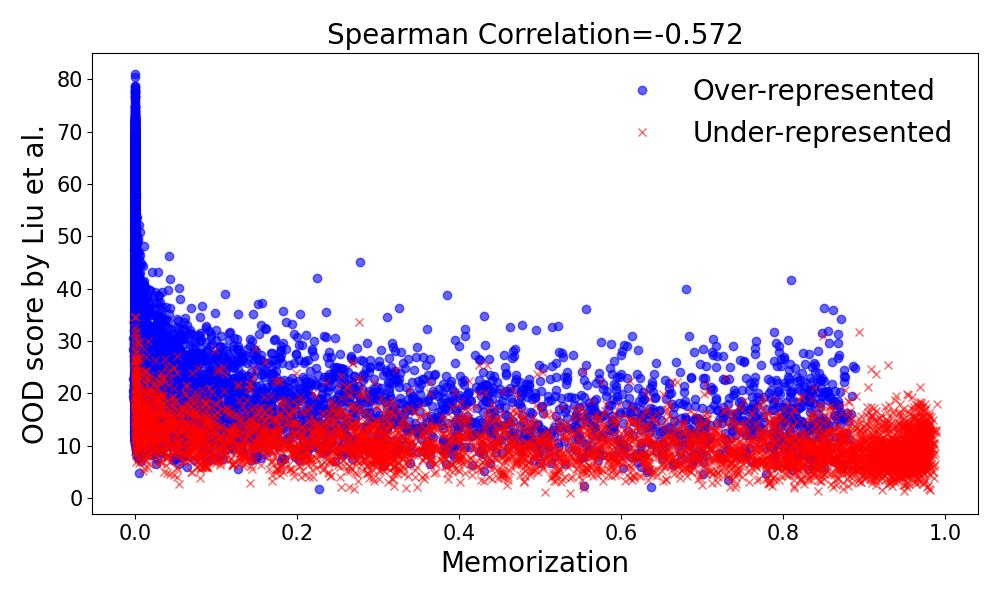}
\caption{MNIST+CIFAR-10}
\includegraphics[width=\textwidth]{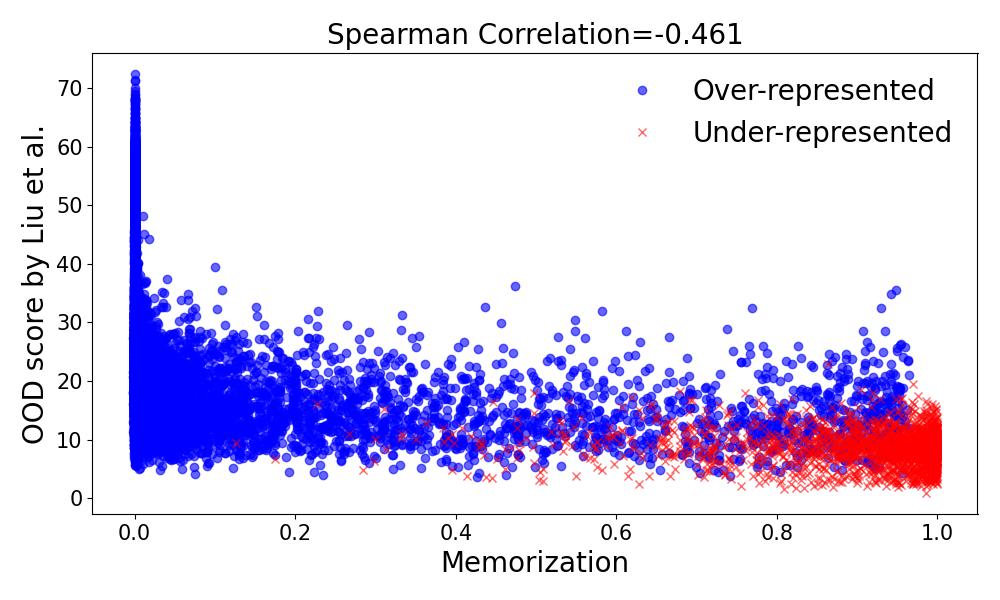}
\caption{MNIST+HM CIFAR-10}
\end{subfigure}
\caption{\textbf{Memorization vs. OOD score by Liu \etal~\cite{liu2020energy}.} Adversary constructs the dataset varying the choice of samples for the under-represented population. ``HM'' refers to Highly Memorized. We observe clearer separation between the populations (blue vs. red) in the second row.}
\label{fig:mem-energy}
\end{figure*}